%% file: iclr2024_conference.tex
\documentclass{article} % For LaTeX2e
\usepackage{iclr2024_conference,times}

\input{math_commands.tex}

\usepackage[utf8]{inputenc} % allow utf-8 input
\usepackage[T1]{fontenc}    % use 8-bit T1 fonts
\usepackage{hyperref}       % hyperlinks
\usepackage{url}            % simple URL typesetting
\usepackage{booktabs}       % professional-quality tables
\usepackage{amsfonts}       % blackboard math symbols
\usepackage{nicefrac}       % compact symbols for 1/2, etc.
\usepackage{microtype}      % microtypography
\usepackage{xcolor}         % colors
\usepackage{algorithm}
\usepackage{algorithmic}
\usepackage[algo2e]{algorithm2e}
\usepackage{graphicx}
\usepackage{wrapfig}
\usepackage{multirow}
\usepackage{enumitem}
\usepackage{algorithm2e}
\usepackage{amsmath}

\usepackage{rotating}
\usepackage{subfigure}
\usepackage{subcaption}

\definecolor{oldlavender}{rgb}{0.47, 0.41, 0.47}
\definecolor{lightsalmon}{rgb}{1.0, 0.63, 0.48}
\definecolor{mediumelectricblue}{rgb}{0.01, 0.31, 0.59}
	\definecolor{purple(munsell)}{rgb}{0.62, 0.0, 0.77}

\title{\texttt{CrysFormer}: Protein Structure Prediction via 3d Patterson Maps and Partial Structure Attention}

\author{%
  Chen Dun* \\
  Rice University \\
  \texttt{cd46@rice.edu} \\
  \And
  Qiutai Pan* \\
  Rice University \\
    \texttt{qp3@rice.edu} \\
  \And
  Shikai Jin \\
  Rice University \\
    \texttt{sj52@rice.edu } \\
\And
  Ria Stevens \\
  Rice University \\
    \texttt{Ria.Stevens@rice.edu} \\
    \And
  Mitchell D. Miller \\
  Rice University \\
    \texttt{mitchm@rice.edu} \\
    \And
   George N. Phillips, Jr. \\
  Rice University \\
    \texttt{georgep@rice.edu} \\
    \And
  Anastasios Kyrillidis \\
  Rice University \\
  \texttt{anastasios@rice.edu} \\
}
\iclrfinalcopy
\begin{document}

\maketitle

\begin{abstract}
Determining the structure of a protein has been a decades-long open question. 
A protein's three-dimensional structure often poses nontrivial computation costs, when classical simulation algorithms are utilized. 
Advances in the transformer neural network architecture --such as AlphaFold2-- achieve significant improvements for this problem, by learning from a large dataset of sequence information and corresponding protein structures. 
Yet, such methods only focus on sequence information; other available prior knowledge, such as protein crystallography and partial structure of amino acids, could be potentially utilized. 
To the best of our knowledge, we propose the first transformer-based model that directly utilizes protein crystallography and partial structure information to predict the electron density maps of proteins. 
Via two new datasets of peptide fragments (2-residue and 15-residue) , we demonstrate our method, dubbed \texttt{CrysFormer}, can achieve accurate predictions, based on a much smaller dataset size and with reduced computation costs. 
\end{abstract}

\def\thefootnote{*}\footnotetext{Authors contributed equally.}\def\thefootnote{\arabic{footnote}}
\input{intro}

\input{definition}

\input{method}

\input{dataset}

\input{results}

\input{discussion}

\clearpage
\bibliography{iclr2024_conference}
\bibliographystyle{iclr2024_conference}

\newpage
\appendix
\section{Appendix}

\input{app_a}

\end{document}

%% file: math_commands.tex
\usepackage{amsmath,amsfonts,bm}

\def\eqref#1{equation~\ref{#1}}
\def\1{\bm{1}}

\DeclareMathAlphabet{\mathsfit}{\encodingdefault}{\sfdefault}{m}{sl}
\SetMathAlphabet{\mathsfit}{bold}{\encodingdefault}{\sfdefault}{bx}{n}

\DeclareMathOperator*{\argmin}{arg\,min}

%% file: intro.tex
\vspace{-0.2cm}\section{Introduction}
\label{sec:introduction}
\vspace{-0.2cm}

Proteins, the biological molecular machines, play a central role in the majority of cellular processes \citep{tanford2004}. 
The investigation of a protein's structure is a classic challenge in biology, given that its function is dictated by its specific conformation. 
Proteins comprise long chains of linked, relatively small organic molecules called \textit{amino acids}, with a set of twenty of them considered as standard. 
However, these underlying polypeptide chains fold into complex three-dimensional structures, as well as into larger assemblies thereof. 
Consequently, biologists aim to establish a standardized approach for experimentally determining and visualizing the overall structure of a protein at a low cost.

In the past decades, there have been three general approaches to the protein structure problem: 
$i)$ ones that rely on physical experimental measurements, such as X-ray crystallography, NMR, or cryo-electron microscopy; see \citep{drenth2007principles} for more details;
$ii)$ protein folding simulation tools based on thermodynamic or kinetic simulation of protein physics \citep{brini2020protein,sippl1990calculation}; and, $iii)$ evolutionary programs based on bioinformatics analysis of the evolutionary history of proteins \citep{vsali1993comparative, roy2010tasser}.
%, homology to previously solved structures and pairwise evolutionary correlations 

Recent advances in machine learning (ML) algorithms have inspired a fourth direction which is to train a deep neural network model on a combination of a large-scale protein structure data set (i.e., the Protein Data Bank \citep{wwpdb2019protein}) and knowledge of the amino acid sequences of a vast number of homologous proteins, to directly predict the protein structure from the protein's  amino acid sequence. 
Recent research projects --such as Alphafold2 \citep{jumper2021highly}-- further show that, with  co-evolutionary bioinformatic information (e.g., multiple sequence alignments), deep learning can achieve highly accurate predictions in most cases. 

% (THIS SECTION NEEDS WORK)
\textbf{Our hypothesis and contributions.} While it is true that computational methods of predicting structures without experimentally confirming data are improving, they are not yet complete --in terms of the types of structures that can be predicted-- and suffer from lack of accuracy in many of the details  \citep{terwilliger2023AlphaFoldNoMatch}. X-ray crystallographic data continues to be a gold standard for critical details describing chemical interactions of proteins. 

Having a robust and accurate way of going directly from an X-ray diffraction pattern to a solved structure would be a strong contribution to the field of X-ray crystallography. 
Such approaches are missing from the literature, with the exception of \cite{pan2023deep}, a recent effort on the same problem based on residual convolutional autoencoders.

Here, we present the first transformer-based model that utilizes protein crystallography and partial structure information to directly predict the electron density maps of proteins, going one step beyond such recent approaches. 
While not yet ready to solve real problems, we demonstrate success on a simplified problem. 
As a highlight, using a new dataset of small peptide fragments of variable unit cell sizes --a byproduct of this work-- we demonstrate that our method, named \texttt{CrysFormer}, can achieve more accurate predictions than state of the art \citep{pan2023deep} with less computations.

Some of our findings and contributions are: \vspace{-0.2cm}
\begin{itemize}[leftmargin=*]  
    \item \texttt{CrysFormer} is able to process the global information in Patterson maps to infer electron density maps; to the best of our knowledge, along with \cite{pan2023deep}, these are the first works to attempt this setting. \vspace{-0.1cm}
    \item \texttt{CrysFormer} can incorporate ``partial structure'' information, when available; we also show that such information could be incorporated in existing solutions that neglected this feature, like the convolutional \texttt{U-Net}-based architectures in \cite{pan2023deep}. However, the \texttt{CrysFormer} architecture still leads to better reconstructions. \vspace{-0.1cm} 
    \item In practice, \texttt{CrysFormer} achieves a significant improvement in prediction accuracy in terms of both Pearson coefficient and mean phase error, while requiring both a smaller number of epochs to converge and less time taken per epoch. \vspace{-0.1cm}
    \item This work introduces a new dataset of variable-cell dipeptide fragments, where all of the input Patterson and output electron density maps were derived from the Protein Databank (PDB) \citep{wwpdb2019protein}, solved by X-ray Crystallography. We will make this dataset publicly available. \vspace{-0.1cm}
\end{itemize}

%% file: definition.tex
\vspace{-0.2cm}\section{Problem Setup and Related Work}
\label{sec:definition}
\vspace{-0.2cm}

\textbf{X-ray crystallography and the crystallographic phase problem.} 
X-ray crystallography has been the most commonly used method to determine a protein's electron density map\footnote{The electron density is a measure of the probability of an electron being present just around a particular point in space; a complete electron density map can be used to obtain a molecular model of the unit cell.} for over 100 years \citep{lattman2008}. 
However, there is an open question, called the crystallographic phase problem, that prevents researchers from utilizing it to predict true structures/electron density maps.

In review, each spot (known as a reflection) in an X-ray crystallography diffraction pattern is denoted by three indices $h, k, l$, known as Miller indices \citep{ashcroft2022solid}. 
These correspond to sets of parallel planes within the protein crystal's unit cell that contribute to producing the reflections.
The set of possible $h, k, l$ values is determined by the radial extent of the observed diffraction pattern.  
Any reflection has an underlying mathematical representation, known as a structure factor, dependent on the locations and scattering factors of all the atoms within the crystal's unit cell. 
In math: \vspace{-0.2cm}
\begin{align}
F(h,k,l) = \sum_{j=1}^{n} f_j \cdot e^{2\pi i(hx_j + ky_j + lz_j)}, \\[-20pt] \nonumber
\end{align}
where the scattering factor and location of atom $j$ are $f_j$ and $(x_j, y_j, z_j)$, respectively. 

A structure factor $F(h,k,l)$ has both an amplitude and a phase component (denoted by $\phi$) and thus can be considered a complex number. 
Furthermore, suppose we knew both components of the structure factors corresponding to all of the reflections within a crystal's diffraction pattern. 
Then, in order to produce an accurate estimate of the electron density at any point $(x,y,z)$ within the crystal's unit cell, we would only need to take a Fourier transform of all of these structures, as in: \vspace{-0.2cm}
\begin{align}
\rho(x,y,z) = \tfrac{1}{V} \cdot \sum_{h,k,l}^{} |F(h,k,l)| \cdot e^{-2\pi i(hx + ky + lz - \phi(h,k,l))}, \\[-20pt] \nonumber
\end{align}
where $V$ is the volume of the unit cell. 
The amplitude $|F(h,k,l)|$ of any structure factor is easy to determine, as it is simply proportional to the square root of the measured intensity of the corresponding reflection. However, it is impossible to directly determine the phase $\phi(h, k, l)$ of a structure factor, and this is what is well-known as the crystallographic phase problem \citep{lattman2008}.

\textbf{Solving the phase problem.} 
Various methods have been developed to solve the crystallography phase problem. 
The three commonly used methods are isomorphous replacement, anomalous scattering, and molecular replacement \citep{lattman2008, jin2020molecular}. 
Also, what is known as direct methods have been successful for small molecules that diffract to atomic resolution, but they rarely work for protein crystallography, due to the difficulty of resolving atoms as separate objects. 
Alternative methods have been developed to solve the phase problem based on intensity measurements alone, known as phase retrieval \citep{guo21, kappeler2017ptychnet, rivenson2018phase}.
However, these methods have not been widely used in X-ray crystallography, because they assume different sampling conditions or were designed for non-crystallographic fields of physics. 
The iterative non-convex Gerchberg–Saxton algorithm \citep{Fienup:82, zalevsky1996gerchberg} is a well-known example of such methods, but requires more measurements than is available in crystallography. 

Although adaptations of the Gerchberg–Saxton algorithm have been proposed for crystallography-like settings, they have not been used to solve the phase problem except in special cases where crystals have very high solvent content \citep{He:mq5029, he2016improving, kingston2022general}.
More recently, 
%Candes et al. 
\cite{candes2013phaselift} introduced the \texttt{Phaselift} method, a convex, complex semidefinite programming approach, and \cite{candes2015phase} the Wirtinger flow algorithm \citep{candes2015phase}, a non-convex phase retrieval method; both these methods have not been applied practically, due to their computationally intensive nature.
%\citep{candes2013phaselift}. 
% Yet, this method has mainly remained theoretical and  
% Similarly, the Wirtinger flow algorithm \citep{candes2015phase}, a non-convex phase retrieval method with theoretical improvements over classic algorithms, has not been widely used in practice.

%% file: method.tex
\vspace{-0.2cm}\section{\texttt{CrysFormer}: Using 3d Maps and Partial Structure Attention}
\label{sec:method}
\vspace{-0.2cm}

%\textbf{A new data-centric solution of X-ray Crystallgraphy} 
Inspired by \citep{hurwitz2020patterson}, we rely on deep learning solutions to directly predict the electron density map of a protein. 
Later in the text, we demonstrate that such a data-centric method achieves both better accuracy and reduced computational cost. 

\textbf{The Patterson function.}
%In order to make an X-ray crystallography dataset more suitable for deep learning, w
We utilize the \textit{Patterson function} \citep{Patterson}, a simplified variation of the Fourier transform from structure factors to electron density, in which all structure factor amplitudes are squared, and all phases are set to zero (i.e., ignored), as in: \vspace{-0.2cm}
\begin{align}
p(u,v,w) = \tfrac{1}{V} \cdot \sum_{h,k,l}^{} |F(h,k,l)|^{2} \cdot e^{-2\pi i(hu + kv + lw)}. \\[-18pt] \nonumber
\end{align}
It is important to note 
%that the Patterson function can be computed without direct access to any phase information of the structure factors. 
%Thus, a 
the Patterson map can be directly obtained from raw diffraction data without the need for additional experiments, or any other information.

Due to the discrete size of the input and output layers in deep learning models, we can discretize and reformulate the electron density map --and its corresponding Patterson map-- as follows: Suppose the electron density map of a molecule in interest is discretized into a $N_1\times N_2 \times N_3$ 3d grid.
The electron density map can then be denoted as $\mathbf{e} \in \mathbb{R}^{N_1 \times N_2 \times N_3}$.
The Patterson map is then formulated as follows, where $\odot$ means matrix element-wise multiplication:
\begin{align*}
    \mathbf{p} = \Re\left(\mathcal{F}^{-1} \left( \mathcal{F}(\mathbf{e}) \odot \mathcal{F}(\widehat{\mathbf{e}})\right) \right) \approx \Re\left(\mathcal{F}^{-1} \left( |\mathcal{F}(\mathbf{e})|^2\right) \right).
\end{align*}
Breaking down the above expression, $\mathcal{F}(\mathbf{e}) \odot \mathcal{F}(\widehat{\mathbf{e}}) \approx |\mathcal{F}(\mathbf{e})|^2$ denotes only the magnitude part of the complex signals, as measured through the Fourier transform of the input signal $\mathbf{e}$. 
Here, $\widehat{\mathbf{e}}$ denotes an inverse-shifted version of $\mathbf{e}$, where its entries follow the shifted rule as in $\widehat{e}_{i, j, k} = e_{N - i, N-j, N-k}$.%\footnote{We note that the Patterson map does not solve the phase problem; it rather ``blindly'' applies the inverse Fourier transform, $\mathcal{F}^{-1}(\cdot)$, and focuses on the real part of the result, as a surrogate estimate of the true electron density signal.
%In an ideal scenario, we would expect that $\mathbf{p} \approx \mathbf{e}$, but this is not the case in general.}

\textbf{Using deep learning.}
We follow a data-centric approach and train a deep learning model, abstractly represented by $g(\boldsymbol{\theta}, \cdot)$, such that given a Patterson map $\mathbf{p}$ as input, it generates an estimate of an electron density map, that resembles closely the true map $\mathbf{e}$. 
Formally, given a data distribution $\mathcal{D}$ and $\left\{\mathbf{p}_i, \mathbf{e}_i\right\}_{i = 1}^n \sim \mathcal{D}$, where $\mathbf{p}_i \in \mathbb{R}^{N_1 \times N_2 \times N_3}$ is the Patterson map that corresponds to the true data electron density map, $\mathbf{e}_{i} \in \mathbb{R}^{N_1 \times N_2 \times N_3}$, deep learning training aims in finding $\boldsymbol{\theta}^\star$ as in: \vspace{-0.1cm}
\begin{align}
\boldsymbol{\theta}^\star &= \underset{\boldsymbol{\theta} }{\argmin} ~\left\{ \mathcal{L}(\boldsymbol{\theta}) := \tfrac{1}{n}\sum_{i = 1}^n \ell(\boldsymbol{\theta};~g, \{\mathbf{p}_i, \mathbf{e}_i\}) = \tfrac{1}{n}\sum_{i = 1}^n \|g(\boldsymbol{\theta}, \mathbf{p}_i) - \mathbf{e}_i\|_2^2 \right\}. \nonumber \\[-10pt] \nonumber
\end{align}
Since we have a regression problem, we use mean squared error as the loss function $\mathcal{L}(\boldsymbol{\theta})$. 

\textbf{Using partial protein structures.} % as prior knowledge.}
Due to the well-studied structure of amino acids, we aim to optionally utilize standardized \textit{partial structures} to aid prediction, when they are available. 
For example, let $\mathbf{u}^j_{i} \in \mathbb{R}^{N_1 \times N_2 \times N_3}$ be the known standalone electron density map of the $j$-th amino acid of the $i$-th protein sample, in a standardized conformation. 
Abstractly, we then aim to optimize: %\vspace{-0.2cm}
\begin{align}
\boldsymbol{\theta}^\star &= \underset{\boldsymbol{\theta} }{\argmin} ~\left\{ \mathcal{L}(\boldsymbol{\theta}) := \tfrac{1}{n}\sum_{i = 1}^n \ell(\boldsymbol{\theta};~g, \{\mathbf{p}_i, \mathbf{e}_i, \mathbf{u}^{j}_{i} \}) = \tfrac{1}{n}\sum_{i = 1}^n \|g(\boldsymbol{\theta}, \mathbf{p}_i,\mathbf{u}^{j}_{i}) - \mathbf{e}_i\|_2^2 \right\}. \nonumber \\[-20pt] \nonumber
\end{align}

\textbf{Challenges and Design Principles.}
We face the difficult learning problem to infer electron density maps $\mathbf{e}$ from Patterson maps $\mathbf{p}$, which involves Fourier transformations. 
\textit{These transformations can be intuitively considered as transforming local information to global information}, which is rare in common deep model use cases. 
Secondly, it is nontrivial to incorporate the partial structure density maps $\mathbf{u}^j_i$ to aid prediction. 
Thirdly, the 3d data format of both our inputs and outputs often increases substantially the computational requirements.
%the input/output dimensions and introduces computational problems in execution. 
Finally, since part of our contributions is novel datasets on this problem, we need to be data efficient due to the expensive dataset creation cost. 
Thus, the main design principles for our model can be summarized as: \vspace{-0.2cm}
\begin{itemize}[leftmargin=*]
  \item \textit{\textcolor{purple}{Design Principle \#1}}: Be able to process the global information in Patterson maps to correctly infer the corresponding electron density maps; \vspace{-0.07cm}
  \item \textit{\textcolor{purple}{Design Principle \#2}}: Be able to incorporate partial structure information, when available; \vspace{-0.07cm}
  \item \textit{\textcolor{purple}{Design Principle \#3}}: Learn to fulfill the above, with reduced computational and data-creation costs. \vspace{-0.5cm}
\end{itemize}

\textbf{Gap in current knowledge.} As an initial attempt, the well-established convolution-based \texttt{U-Net} model \citep{ronneberger2015u} could be utilized for this task.
This is the path followed in \citep{pan2023deep}.
However, classical \texttt{U-Nets} cannot fulfill the design principles above, since: 
$i)$ they mostly rely on local information within CNN layers; such a setup is not suitable when Patterson maps are available, since the latter do not have meaningful local structures. 
$ii)$ It is not clear (or, at best, non-trivial) to incorporate any partial protein structures prior information, since the latter is in a different representation domain, compared to Patterson maps. 
Finally, $iii)$ a large 3d \texttt{U-Net} model is computationally expensive and inefficient, due to the 3d filter convolution computation. 

\textbf{Our proposal: \texttt{CrysFormer}}.
We propose \texttt{CrysFormer}, a novel, 3d Transformer model \citep{vaswani2017attention, chen21} with a new self-attention mechanism to process Patterson maps and partial protein structures, to directly infer electron density maps with reduced costs.

%Focusing on the design principles above, for \textit{\textcolor{purple}{design principle \#1}}, 
Inspired by recent research on the potential connection between Fourier transforms and the self-attention mechanism, found in the Transformer model \citep{leethorp2022fnet}, \texttt{CrysFormer} captures the global information in Patterson maps and ``translates'' it into correct electron density map predictions, via our proposed self-attention mechanism (\textit{\textcolor{purple}{Design Principle \#1}}). 
\texttt{CrysFormer} does not need an encoder-decoder structure \citep{vaswani2017attention} and artificial information bottlenecks \citep{cheng19} --as in the \texttt{U-Net} architecture-- to force the learning of global information. %, which might cause a loss of finer details in the final electron density prediction. 

By definition, \texttt{CrysFormer} is able to handle additional partial structure information, which comes from a different domain than the Patterson maps (\textit{\textcolor{purple}{Design Principle \#2}}; more details below). 

%For \textbf{\textcolor{purple}{Aim \#3}}, inspired by recent research in 3D visual Transformers, we propose that 
Finally, by using efficient self-attention between 3d image patches, we can significantly reduce the overall computation cost. 
Detaching our model from an encoder-decoder architecture further reduces the required depth of the model and, thus, the overall training cost (\textit{\textcolor{purple}{Design Principle \#3}}).

\begin{figure*}[!htp]
    \vspace{-0.0cm}
    \centering
    \includegraphics[width=0.95\linewidth]{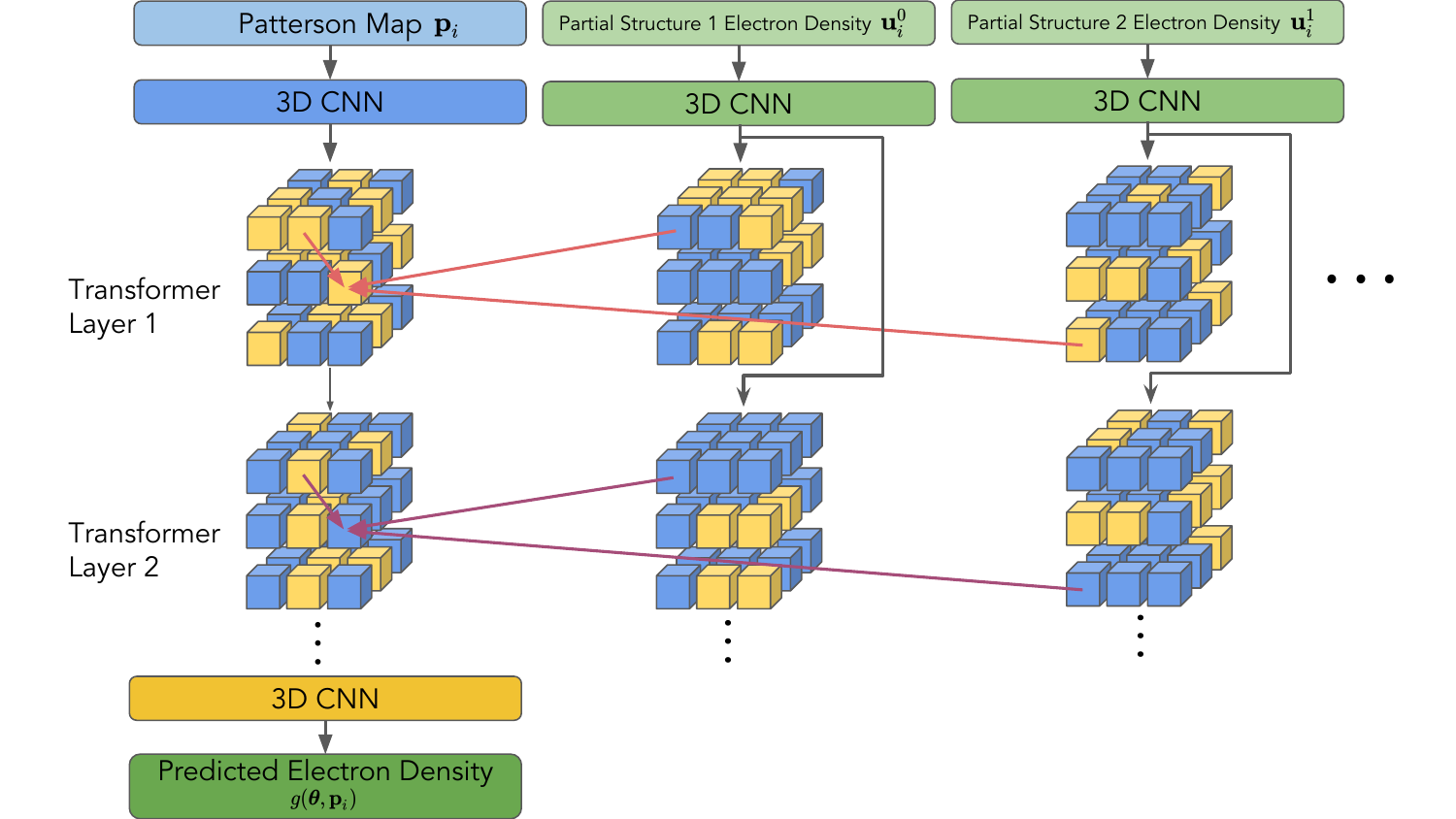} \vspace{-0.1cm}
    \caption{\small Abstract depiction of the \texttt{Crysformer}, which utilizes a one-way attention mechanism (\textcolor{red}{red} and \textcolor{purple}{purple} arrows) to incorporate the partial structure information. The tokens from the additional partial structure all come from initial 3d CNN embedding and are not passed to the next layer.} %\textcolor{red}{The figure can be squeezed vertically a bit (last layer can go slightly up to save white space). Also, Chen, please use shapes like the ones we use the Transist figures. Use the same thickness for all arrows. Use math symbols: e.g., Next to patterson map box, we should have $\mathbf{p}$; next to structure 1 we should have $\mathbf{u}^1$, etc. It is not initially clear how the information flows (from top to bottom); we could also use $\textbf{(a)}, \textbf{(b)}, \dots $ to connect with the text. Finally, if we have more layer depth or more structures, this should be indicated with three dots.}}
    \label{fig:crysformer}
    \vspace{-0.3cm}
\end{figure*}

\textbf{The architecture of the \texttt{CrysFormer}.} 
We follow ideas of a 3d visual Transformer \citep{chen21} by partitioning the whole input 3d Patterson map $\mathbf{p}_i \in \mathbb{R}^{N_1 \times N_2 \times N_3}$ input into a set of smaller 3d patches. 
We embed them into one-dimensional ``word tokens'', and feed them into a multi-layer, encoder-only Transformer module. 
If partial structures $\mathbf{u}^j_i$ are also available, we will partition them into 3d patches and embed them into additional tokens that are sent to each self-attention layer.
This way, the tokens in each layer can also ``attend'' the election density of partial structures, as a reference for final global electron density map predictions. Finally, we utilize a 3d convolutional layer to transform ``word-tokens'' back into a 3d electron density map.\footnote{We also utilize 3d convolutional layer(s) at the very beginning of the execution to expand the number of channels of the Patterson map (and potentially partial structure) inputs.} 
See Figure \ref{fig:crysformer}.

Mathematically, we report the following: 
The first part is the preprocessing and partitioning of input Patterson maps $\mathbf{p}$ and additional partial structures $\mathbf{u}^j$ into 3d patches of size $d_1\times d_2 \times d_3$. 
We embed those patches into one-dimensional tokens with dimension $d_t$, using of a small MLP, and add them with a learned positional embedding; this holds for both Patterson maps and structures, as below:
\vspace{-0.0cm}

\begin{minipage}[h]{0.5\textwidth}
\vspace{0pt}
\begin{center}
    \textbf{Patterson maps} $\mathbf{p}$
\end{center}
\begin{small}
\begin{align*}
    \mathbf{X}^0 &=\texttt{3DCNN}_{\mathbf{W}_c}(\mathbf{p}) \in \mathbb{R}^{c \times N_1 \times N_2 \times N_3}\\
    \mathbf{X}^0 &= \texttt{Partition}(\mathbf{X}^0) \in \mathbb{R}^{ \frac{N_1}{d_1} \times \frac{N_2}{d_2} \times \frac{N_3}{d_3} \times (cd_1d_2d_3)} \\
    \mathbf{X}^0 &= \texttt{Flatten}(\mathbf{X}^0) \in \mathbb{R}^{ \frac{N_1N_2N_3}{d_1d_2d_3} \times (cd_1d_2d_3)} \\
    \mathbf{X}^0 &= \texttt{MLP}_{\mathbf{W}_c}(\mathbf{X}^0) \in \mathbb{R}^{\frac{N_1N_2N_3}{d_1d_2d_3} \times d_t} \\
    \mathbf{X}^0 &= \mathbf{X}^0 + \texttt{PosEmbedding}(\tfrac{N_1N_2N_3}{d_1d_2d_3})
\end{align*}
\end{small}
\end{minipage}
\hfill
\begin{minipage}[h]{0.5\textwidth}
\vspace{0cm}
\begin{center}
    \textbf{Partial structures} $\mathbf{u}^j$
\end{center}
\vspace{0.3cm}
\begin{tabular}{|p{\textwidth}}
\begin{small}
{$\!\begin{aligned}
    \mathbf{U}^j &=\texttt{3DCNN}_{\mathbf{W}_p}(\mathbf{u}^j) \in \mathbb{R}^{c \times N_1 \times N_2 \times N_3}\\
    \mathbf{U}^j &= \texttt{Partition}(\mathbf{U}^j) \in \mathbb{R}^{ \frac{N_1}{d_1} \times \frac{N_2}{d_2} \times \frac{N_3}{d_3} \times (cd_1d_2d_3)} \\
    \mathbf{U}^j &= \texttt{Flatten}(\mathbf{U}^j) \in \mathbb{R}^{ \frac{N_1N_2N_3}{d_1d_2d_3} \times (cd_1d_2d_3)} \\
    \mathbf{U}^j &= \texttt{MLP}_{\mathbf{W}_p}(\mathbf{U}^j) \in \mathbb{R}^{ \tfrac{N_1N_2N_3}{d_1d_2d_3} \times d_t} \\
    \mathbf{U}^j &= \mathbf{U}^j + \texttt{PosEmbedding}(\tfrac{N_1N_2N_3}{d_1d_2d_3})
\end{aligned}$}
\end{small}
\end{tabular}
\end{minipage}

% \begin{small}
% \begin{align*}
%     \mathbf{X}^0 &=\texttt{3DCNN}_{\mathbf{W}_c}(\mathbf{p}) \in \mathbb{R}^{c \times N_1 \times N_2 \times N_3}\\
%     \mathbf{X}^0 &= \texttt{Partition}(\mathbf{X}^0) \in \mathbb{R}^{ \frac{N_1}{d_1} \times \frac{N_2}{d_2} \times \frac{N_3}{d_3} \times (cd_1d_2d_3)} \\
%     \mathbf{X}^0 &= \texttt{Flatten}(\mathbf{X}^0) \in \mathbb{R}^{ \frac{N_1N_2N_3}{d_1d_2d_3} \times (cd_1d_2d_3)} \\
%     \mathbf{X}^0 &= \texttt{MLP}_{\mathbf{W}_c}(\mathbf{X}^0) \in \mathbb{R}^{\frac{N_1N_2N_3}{d_1d_2d_3} \times d_t} \\
%     \mathbf{X}^0 &= \mathbf{X}^0 + \texttt{PosEmbedding}(\tfrac{N_1N_2N_3}{d_1d_2d_3}) \\
%     \mathbf{U}^j &=\texttt{3DCNN}_{\mathbf{W}_p}(\mathbf{u}^j) \in \mathbb{R}^{c \times N_1 \times N_2 \times N_3}\\
%     \mathbf{U}^j &= \texttt{Partition}(\mathbf{U}^j) \in \mathbb{R}^{ \frac{N_1}{d_1} \times \frac{N_2}{d_2} \times \frac{N_3}{d_3} \times (cd_1d_2d_3)} \\
%     \mathbf{U}^j &= \texttt{Flatten}(\mathbf{U}^j) \in \mathbb{R}^{ \frac{N_1N_2N_3}{d_1d_2d_3} \times (cd_1d_2d_3)} \\
%     \mathbf{U}^j &= \texttt{MLP}_{\mathbf{W}_p}(\mathbf{U}^j) \in \mathbb{R}^{ \tfrac{N_1N_2N_3}{d_1d_2d_3} \times d_t} \\
%     \mathbf{U}^j &= \mathbf{U}^j + \texttt{PosEmbedding}(\frac{N_1N_2N_3}{d_1d_2d_3}) \\[-20pt] 
% \end{align*} 
% \end{small}
As shown in Figure \ref{fig:crysformer}, we design an efficient attention mechanism such that $i)$ only tokens from Patterson maps attend tokens from the partial structures; $ii)$ the tokens from the additional partial structures are not passed to the next layer. 
This is based on that the partial structure electron density information should be used by the model as a stable reference to attend to in each layer.  

This one-way attention also greatly reduces the overall communication cost.
In particular, let the token sequence length be $S=\tfrac{N_1N_2N_3}{d_1d_2d_3}$ and let $d_{h}$ denote the dimension of the attention head. Assuming we have $H$ attention heads and $L$ layers, \texttt{CrysFormer} uses the following attention mechanism: \vspace{-0.05cm}
\begin{align*}
    \mathbf{U} &= \texttt{Concat}_{j=1}^{J}(\mathbf{U}^j) \in \mathbb{R}^{ (SJ) \times d_t} \\
    \mathbf{A}^h &=\texttt{Softmax}\left((\mathbf{W}^h_q\mathbf{X}^{\ell})^{\top}(\texttt{Concat}(\mathbf{W}^h_k\mathbf{X}^{\ell},\mathbf{W}^h_{k'}\mathbf{U})\right) \in \mathbb{R}^{S \times (S+SJ)}; \\
    \mathbf{\widehat{V}}^h &= \mathbf{A}^h \left(\texttt{Concat}(\mathbf{W}^h_v \mathbf{X}^{\ell},\mathbf{W}^h_{v'} \mathbf{U})\right) \in \mathbb{R}^{S \times d_{h}}; \\
    \mathbf{O} &= \mathbf{W}_o\texttt{Concat}\left(\mathbf{\widehat{V}}^0,~\mathbf{\widehat{V}}^1, \dots, ~\mathbf{\widehat{V}}^H \right) \in \mathbb{R}^{S \times d_{t}}; \\
    \mathbf{X}^{\ell+1} &= \mathbf{W}_{\text{ff2}}(\texttt{ReLU}(\mathbf{W}_{\text{ff1}}\mathbf{O})),
    \\[-10pt]
\end{align*} 
where, omitting the layer index, $\mathbf{W}^h_q$, $\mathbf{W}^h_k$, $\mathbf{W}^h_v$ are the trainable query, key, and value projection matrices of the $h$-th attention head for tokens from the Patterson map, and $\mathbf{W}^h_{k'}$, $\mathbf{W}^h_{v'}$ are the corresponding matrices for tokens from the partial structure, each with dimension $d_{h}$.  Further, $\mathbf{W}_{\text{ff1}}$ and $\mathbf{W}_{\text{ff2}}$ are the trainable parameters of the fully-connected layers. 
We omit skip connections and layer normalization modules just to simplify notation, but these are included in practice. 

As a final step, we transform the output embedding back to a 3d electron density map, as follows:
\begin{align*}
   g(\boldsymbol{\theta}, \mathbf{p}) &= \texttt{tanh}(\texttt{3DCNN}_{\mathbf{W}_o}(\texttt{Rearrange}(\texttt{MLP}(\mathbf{X}^{L})))) \in \mathbb{R}^{N_1 \times N_2 \times N_3},
\end{align*}
and, as stated previously, we use as our loss function the standard mean squared error loss.

%% file: dataset.tex
\vspace{-0.2cm}\section{New Datasets}
\label{sec:dataset}
\vspace{-0.2cm}

We generate datasets of protein fragments, where input Patterson and output electron density maps are derived from Protein Databank (PDB) entries of proteins solved by X-ray Crystallography \citep{wwpdb2019protein}. 
We start from a curated basis of $\sim 24,000$ such protein structures. %, which were curated based on several criteria such as sequence length. 
Then from a random subset of about half of these structures, we randomly select and store segments of adjacent amino acid residues.  
These examples are consisted of dipeptides (two residues) and 15-residues, leading to two datasets that we introduce with this work. 
The latter dataset contains 15 residues, where at most 3 residues could be shared between different examples. 
Using the \texttt{pdbfixer} Python API \citep{eastman2017openmm}, we remove all examples that either contain nonstandard residues or have missing atoms from our initial set. We also apply a few standardized modifications. %We apply a short molecular dynamics simulation to each of the examples to remove any potential clashes or irregular conformations, as well as a set of standardized modifications.

For our dipeptide dataset, we then iteratively expand the unit cell dimensions for each example, starting from the raw $\max-\min$ ranges in each of the three axis directions, attempting to create a minimal-size unit cell where the minimum atomic contact is at least $2.75$ Angstroms ({\AA}).\footnote{An Angstrom is a metric unit of length equal to $10^{-10}$m.} For our 15-residue dataset, we instead place atoms in fixed unit cells of size $41$ {\AA} x $30$ {\AA} x $24$ {\AA} to simplify the now much harder problem.
After this, all examples that still contain atomic contacts of less than $2.75$ {\AA} are discarded.
The examples are then reoriented via a reindexing operation, such that the first axis is always the longest and the third axis is always the shortest.  

One issue leading to potential ambiguity in interpreting Patterson maps is their invariance to translation of the entire corresponding electron density \citep{hurwitz2020patterson}.
To tackle this, we center all atomic coordinates such that the center of mass is in the center of the corresponding unit cell. 
This means that our model's predicted electron densities would always be more or less centered in the unit cell. 
We note that this is also the case for the majority of actual protein crystals. 

Structure factors for each remaining example, as well as those for the corresponding partial structures for each of the present amino acids, are generated using the \texttt{gemmi sfcalc} program \citep{wojdyr2022gemmi} to a resolution of $1.5$ {\AA}. 
An electron density and Patterson map for each example are then obtained from those structure factors with the \texttt{fft} program of the \texttt{CCP4} program suite \citep{read1988phased, Winn:dz5219}; partial structure densities are obtained in the same manner.
We specify a grid oversampling factor of $3.0$, resulting in a $0.5$ {\AA} grid spacing in the produced maps. 
All these maps are then converted into PyTorch tensors. 
We then normalize the values in each of the tensors to be in the range $[-1, ~1]$.
Since, in our PyTorch implementation, all examples within a training batch are of the same size, we remove all examples from the tensor-size bins containing fewer examples than a specified minimum batch size.%, after determining the overall maximum and minimum values present in each.

%\textcolor{red}{This part needs refinement.}
%We have also generated a dataset of fragments of 15 adjacent residues, also derived from the $\sim 24,000$ curated PDB entries.  We enforced that at most 3 residues could be shared between different examples.  The rest of the data generation process was the same as for generating dipeptide examples, except that we placed atoms in fixed unit cell sizes of $41$ {\AA} x $30$ {\AA} x $24$ {\AA} to simplify the now much harder problem initially.

%% file: results.tex
\vspace{-0.2cm}\section{Experiments}
\label{sec:results}
\vspace{-0.2cm}

\textbf{Baselines.} 
There are no readily available off-the-self solutions for our setting, as our work is one of the first of this kind. 
As our baseline, we use a CNN-based \texttt{U-Net} model \citep{pan2023deep}; this architecture is widely used in image transformation tasks \citep{ronneberger2015u, yan21}.  

%In particular, our baseline \texttt{U-Net} architecture is an extended version of the architecture presented in previous work \citep{pan2023deep}.  
For comparison, we have further enhanced this vanilla \texttt{U-Net} with $i)$ additional input channels to incorporate the partial structure information, despite being evidently unsound; and $ii)$ a refining model procedure, which retrains the \texttt{U-Net} using previous model predictions as additional input channels. 
Both of these extensions are shown to greatly improve the performance of the vanilla \texttt{U-Net}. 
We refer the reader to the appendix for more details on our baseline model architecture. 

\textbf{Metrics.} 
During testing, we calculate the Pearson correlation coefficient between the ground truth targets $\mathbf{e}$ and model predictions $g(\boldsymbol{\theta}, \mathbf{p})$; the larger this coefficient is, the better. 
Let us denote a model prediction as $\mathbf{e'}$.  We define $\bar{\mathbf{e}}=\frac{1}{N_1N_2N_3}\sum_{i,j,k}\mathbf{e}_{i,j,k}$ and $\bar{\mathbf{e}}'=\frac{1}{N_1N_2N_3}\sum_{i,j,k}\mathbf{e}'_{i,j,k}$.  Then, the Pearson correlation coefficient between $\mathbf{e}$ and $\mathbf{e'}$ is as below:
\begin{equation}
\texttt{PC}(\mathbf{e},\mathbf{e}')=\frac{\sum_{i,j,k = 1}^{N_1,N_2,N_3} (\mathbf{e}'_{i,j,k} - \bar{\mathbf{e}}') (\mathbf{e}_{i,j,k} - \bar{\mathbf{e}})}{\rule{0pt}{1.6em} \sqrt{\sum_{i,j,k = 1}^{N_1,N_2,N_3} (\mathbf{e}'_{i,j,k} - \bar{\mathbf{e}}') + \epsilon} \cdot \: \sqrt{\sum_{i,j,k = 1}^{N_1,N_2,N_3} (\mathbf{e}_{i,j,k} - \bar{\mathbf{e}}) + \epsilon}},
\end{equation}
where $\epsilon$ is a small constant to prevent division by zero.
To demonstrate how well our methods solve the phase problem, we also perform phase error analysis on our models' final post-training predictions using the \texttt{cphasematch} program of the \texttt{CCP4} program suite \citep{cphasematch}.  
We report the mean phase errors of our predictions in degrees, as reported by \texttt{cphasematch}, where a smaller phase error is desirable. Finally, we compare the convergence speed and computation cost of both methods.

%GPU memory: 2097, 2097, 2099, 3059
\begin{table*}[!htp]
\centering
\begin{footnotesize}
\begin{tabular}{ccccc}
\toprule
    Method  & Mean $\texttt{PC}(\mathbf{e},\mathbf{e}')$ & Mean Phase Error & Epochs & Time per epoch (mins.) \\ \midrule
    \texttt{U-Net} \citep{pan2023deep} & 0.735 & 67.40$^{\circ}$ & 50 & 28.93  \\
    \texttt{U-Net+R} (This work) & 0.775 & 58.67$^{\circ}$ &  90 & 29.06  \\
    \texttt{U-Net+PS+R} (This work) & 0.839 & 51.34$^{\circ}$ &  90 & 29.31  \\
    \texttt{CrysFormer} (This work) & \textcolor{teal}{\textbf{0.939}} & \textcolor{teal}{\textbf{35.16}$^{\circ}$} & \textcolor{teal}{\textbf{35}} & \textcolor{teal}{\textbf{12.37}} \\
    \bottomrule
\end{tabular}
\end{footnotesize}
\caption{\small \texttt{CrysFormer} versus baselines on the dipeptide dataset. \texttt{U-Net+R} refers to adding the refining procedure to \texttt{U-Net} training; \texttt{U-Net+PS+R} refers to adding further partial structures as additional channels. } 
\vspace{-0.3cm}
\label{text_classification}
\end{table*}

\textbf{Results on two-residues.}
A summary of our results on our dipeptide dataset, which consisted of $1,894,984$ training and $210,487$ test cases, is provided in Table \ref{text_classification}.
Overall, \texttt{CrysFormer} achieves a significant improvement in prediction accuracy in terms of both the Pearson coefficient and phase error, while requiring a shorter time (in epochs) to converge. 
\texttt{CrysFormer} also incurs much less computation cost which results in significantly reduced wall clock time per epoch. % \footnote{as measured by Python's $time.perf\_counter$}.  

\begin{figure}[!htp]
\vspace{-0.5cm}
\subfigure[\texttt{U-Net+R}]{\includegraphics[width=0.3\textwidth, height= 3.5cm]{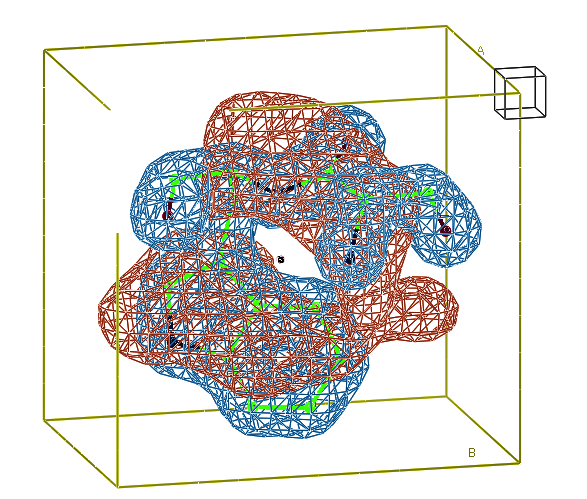}} \hfill
\subfigure[\texttt{U-Net+PS+R}]{\includegraphics[width=0.3\textwidth, height= 3.5cm]{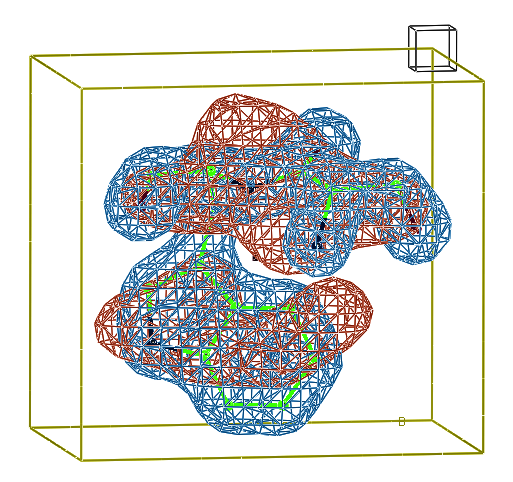}} \hfill
\subfigure[\texttt{CrysFormer}]{\includegraphics[width=0.3\textwidth, height= 3.5cm]{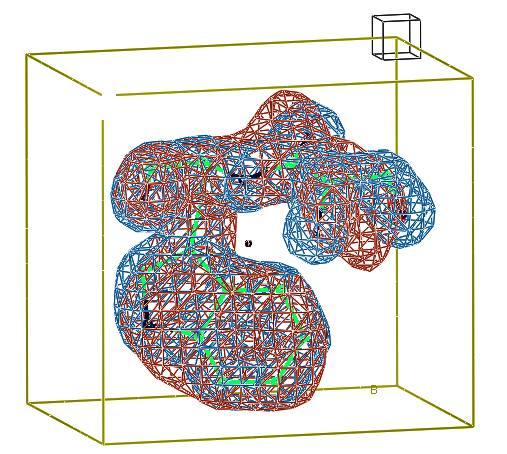}} \hfill
\vspace{-0.35cm}
\caption*{Serine + Tryptophan} \label{fig:ST}
%\subfigure[\texttt{U-Net+R}]{\includegraphics[width=0.3\textwidth, height= 3.4cm]{images/4BSX_1.pd_88_unet_recycling2-IRM.png}} \hfill
%\subfigure[\texttt{U-Net+PS+R}]{\includegraphics[width=0.3\textwidth, height= 3.4cm]{images/4BSX_1_pd_88_unet_channels_recycling2-IRM.png}} \hfill
%\subfigure[\texttt{CrysFormer}]{\includegraphics[width=0.3\textwidth, height= 3.4cm]{images/4BSX_1.pd_88-IRM.png}} \hfill
%\vspace{-0.35cm}
%\caption{Aspartic Acid + Valine} \label{fig:AV}
%\subfigure[\texttt{U-Net+R}]{\includegraphics[width=0.3\textwidth, height= 3.5cm]{images/6FJK_1.pd_315_unet_recycling2-IRM.png}} \hfill
%\subfigure[\texttt{U-Net+PS+R}]{\includegraphics[width=0.3\textwidth, height= 3.5cm]{images/6FJK_1.pd_315_unet_channels_recycling2-IRM.png}} \hfill
%\subfigure[\texttt{CrysFormer}]{\includegraphics[width=0.3\textwidth, height= 3.5cm]{images/6FJK_1.pd_315-IRM.png}} \hfill
%\vspace{-0.35cm}
%\caption{Aspartic Acid + Lysine} \label{fig:AL}
\subfigure[\texttt{U-Net+R}]{\includegraphics[width=0.3\textwidth, height= 3.5cm]{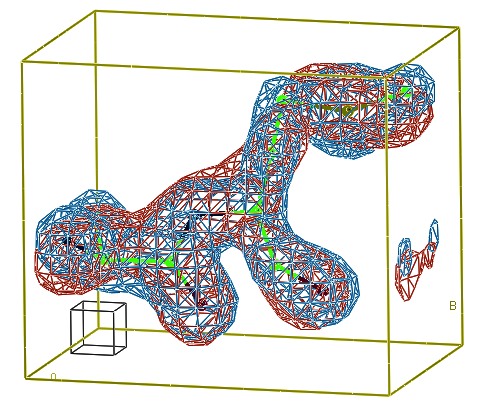}} \hfill
\subfigure[\texttt{U-Net+PS+R}]{\includegraphics[width=0.3\textwidth, height= 3.5cm]{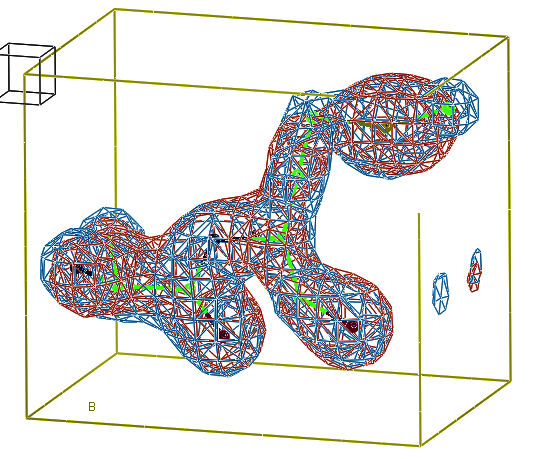}} \hfill
\subfigure[\texttt{CrysFormer}]{\includegraphics[width=0.3\textwidth, height= 3.5cm]{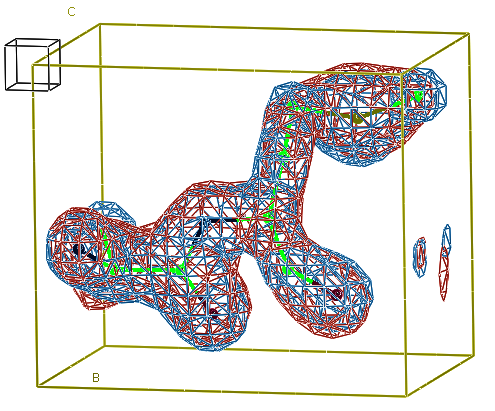}} \hfill
\vspace{-0.3cm}
\caption*{Alanine + Methionine} \label{fig:AM}
\vspace{-0.2cm}
\caption{\small Visualization of electron density predictions for baselines and \texttt{CrysFormer}: Ground truth density maps are shown in blue, while predictions are shown in red. The model used to generate the ground truth electron density is shown in stick representation for reference.}
\label{fig:main} \vspace{-0.3cm}
\end{figure}

We further visualize some of the predictions in Figure \ref{fig:main}, comparing side by side those made by the baselines and the \texttt{CrysFormer}. 
\texttt{CrysFormer} produces more accurate predictions in terms of both global and local structures. 
This verifies our hypothesis that $i)$ the self-attention mechanism can better capture the global information in Patterson maps, and $ii)$ the removal of the \texttt{U-Net}'s encoder-decoder structure prevents loss of information and improves the reproduction of finer details. % in the final electron density prediction. 

E.g., the top row of Figure \ref{fig:main} represents a class of examples containing a large aromatic residue, Tryptophan. 
\texttt{U-Net+R} models consistently produce poor predictions in this case, while the \texttt{CrysFormer} better handles such residues.
\texttt{U-Net+PS+R} shows that both providing additional input channels and using the refining procedure improves results even for \texttt{U-Net} architectures; yet, \texttt{CrysFormer} still provides better reconstruction.
%\textcolor{teal}{(although test examples with aromatic residues still tended to have below average prediction quality)}.
%Figure 3 represents an example in which the additional partial structure input channels provided to the U-Net provided a substantial increase in prediction quality, allowing it to produce a prediction similar to that of the \texttt{CrysFormer}.  
%Meanwhile, Figure 4 represents an example in which both providing additional input channels to the U-Net and switching to \texttt{CrysFormer} provided noticeable improvements in prediction quality.
%Figures \ref{fig:AV}, \ref{fig:AL} and \ref{fig:AM} show similar results for the cases of Aspartic Acid + Valine, Aspartic Acid + Lysine and Alanine + Methionine, respectively.
More visualizations can be found in the appendix.
%\textcolor{red}{I would suggest we move some of those in the appendix to save space.}
%Figure 5 represents an example in which all of our models provided a reasonably accurate prediction; this was actually the case for the majority of the test set examples.

\begin{figure}[!ht]
\vspace{-0.4cm}
\centering
\subfigure[\empty]{\includegraphics[width=0.44\textwidth]{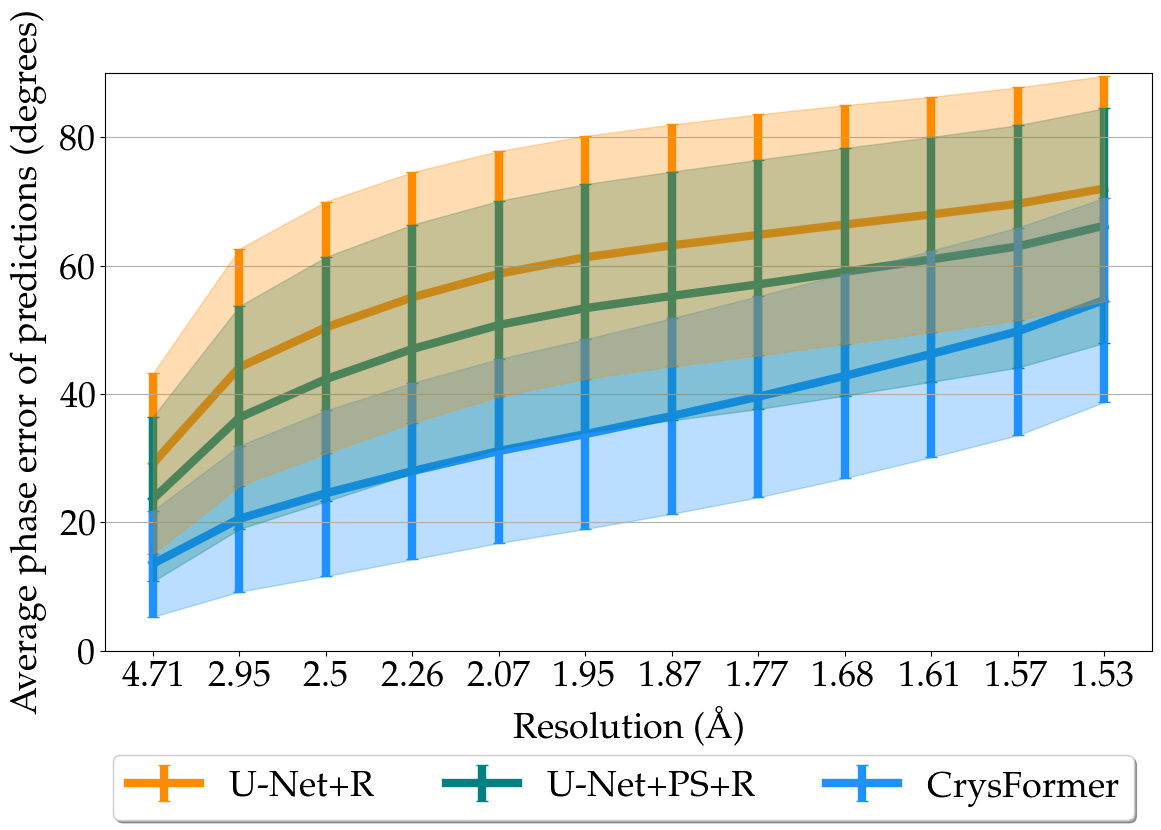}} \hspace{0.5cm}
\subfigure[\empty]{\includegraphics[width=0.44\textwidth]{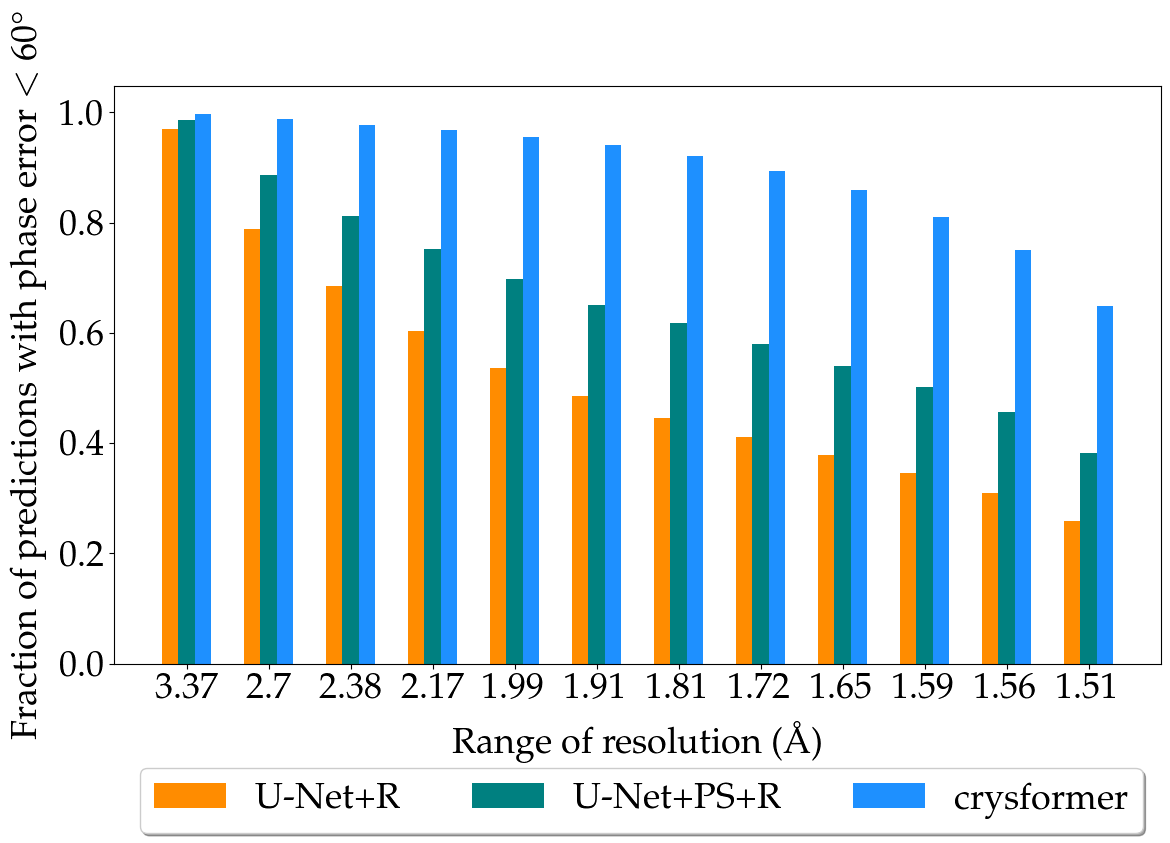}} \hfill
\vspace{-0.2cm}
\caption{\small Dipeptide dataset. \textbf{Left}: Average phase error of model predictions against reflection resolution. \textbf{Right}:  Fraction of model predictions for which phase error is $< 60^{\circ}$ at various ranges of resolution.} 
\label{phase_error} \vspace{-0.3cm}
\end{figure}

We further plot the calculated average mean phase errors of the predictions of our models against reflection resolution, see left panel of Figure \ref{phase_error}. 
The predictions made by \texttt{CrysFormer} have lower mean phase error, compared to baselines.  
This means that the \texttt{CrysFormer} predictions, on average, can reproduce better the general shape, as well as finer details of the ground truth electron densities. % better than the predictions made by the U-Net architectures. 

Finally, we generate a chart of the fraction of our models' predictions for which the calculated mean phase error is $< 60^{\circ}$ at various ranges of resolution. 
We consider such predictions to accurately reproduce the level of detail specified by that resolution range. 
This is shown on the right panel in Figure \ref{phase_error}.  
At all resolution ranges, \texttt{CrysFormer} predictions are clearly better than that of the \texttt{U-Net}-based models.  In particular, for \texttt{CrysFormer}, we still have a majority of predictions with phase error $< 60^{\circ}$ even at the highest ranges of resolution.

\textbf{Results on 15-residues.}
On our dataset of 15-residue examples, which consisted of only $165,858$ training and $16,230$ test cases (less than one-tenth the size of our dipeptide dataset), we trained for 80 epochs to a final average test set Pearson correlation of about $0.747$.  We then performed a refining training run of 20 epochs, incorporating the original training run's predictions as additional input channels when training the \texttt{CrysFormer}, and obtained an improved average test set Pearson correlation of about $0.77$ and phase error of about $67.66$. On both of these runs, we used the Nyström approximate attention mechanism \citep{Xiong_Zeng_Chakraborty_Tan_Fung_Li_Singh_2021} when incorporating our partial structure information to reduce time and space costs.  Even still, each training epoch still took about $6.28$ hours to complete.  Thus due to time considerations, we decided not to attempt to train a U-Net on this dataset for purposes of comparison.
%\textcolor{red}{We should state the reason why we test only Crysformer on this dataset -- I assume time, but we need to make it clear here.}

\begin{figure}[!htp]
\vspace{-0.5cm}
\centering
\subfigure[\empty]{\includegraphics[width=0.44\textwidth]{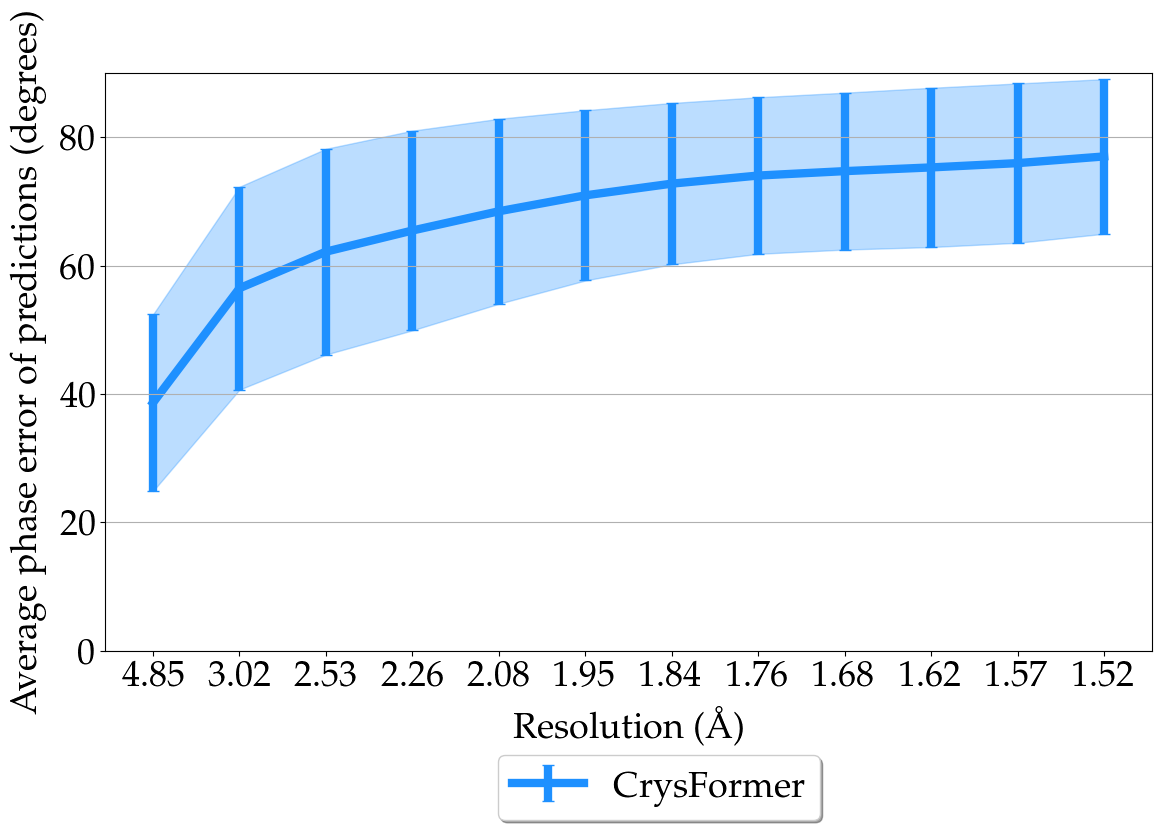}} \hspace{0.5cm}
\subfigure[\empty]{\includegraphics[width=0.44\textwidth]{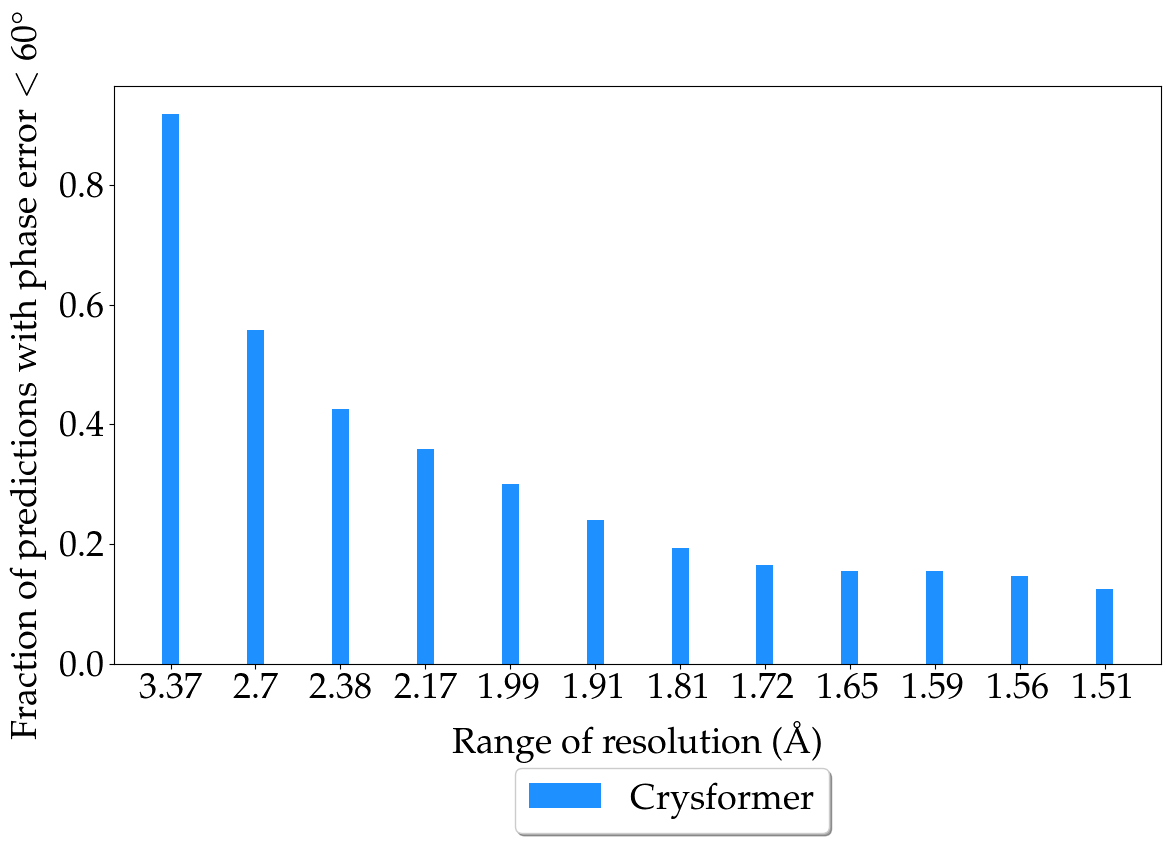}} \hfill
\vspace{-0.3cm}
\caption{\small \textbf{Left}: Average phase error of model predictions on 15-residue dataset against reflection resolution. \textbf{Right}:  Fraction of model predictions on 15-residue dataset for which phase error is $< 60^{\circ}$ at various ranges of resolution.} 
\label{phase_error15} \vspace{-0.6cm}
\end{figure}

We provide visualizations of some model predictions in Figure \ref{pred15}; more can again be found in the appendix.  We also plot the average mean phase errors of the predictions of our models against reflection resolution, as well as the fraction of our models' predictions for which the calculated mean phase error is $< 60^{\circ}$ at various ranges of resolution in Figure \ref{phase_error15}.  These results show that this is a more difficult dataset with reduced sample size; yet \texttt{CrysFormer} predictions tend to accurately reproduce details of the desired electron densities.  
%Since our examples were now large enough for us to start applying crystallographic refinement procedures to them, we decided that this was not a significant issue. 

%\begin{figure}[!htp]
\begin{wrapfigure}{r}{0.5\textwidth}
\vspace{-0.5cm}
\centering
\subfigure[\texttt{5E4V$\_$1$\_$25}]{\includegraphics[width=0.44\linewidth]{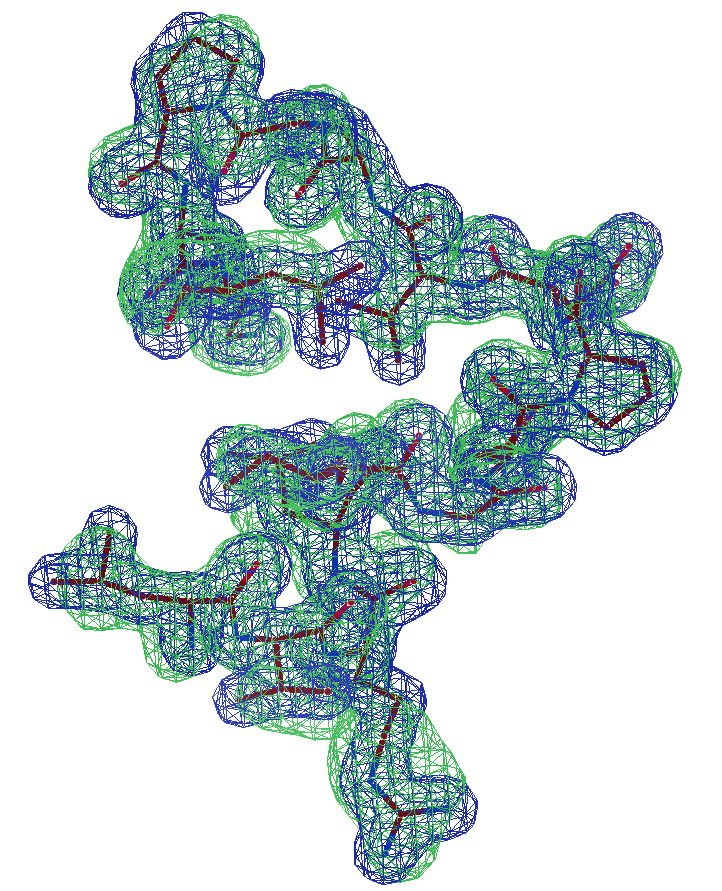}}
\subfigure[\texttt{3HN7$\_$1$\_$133}]{\includegraphics[width=0.36\linewidth]{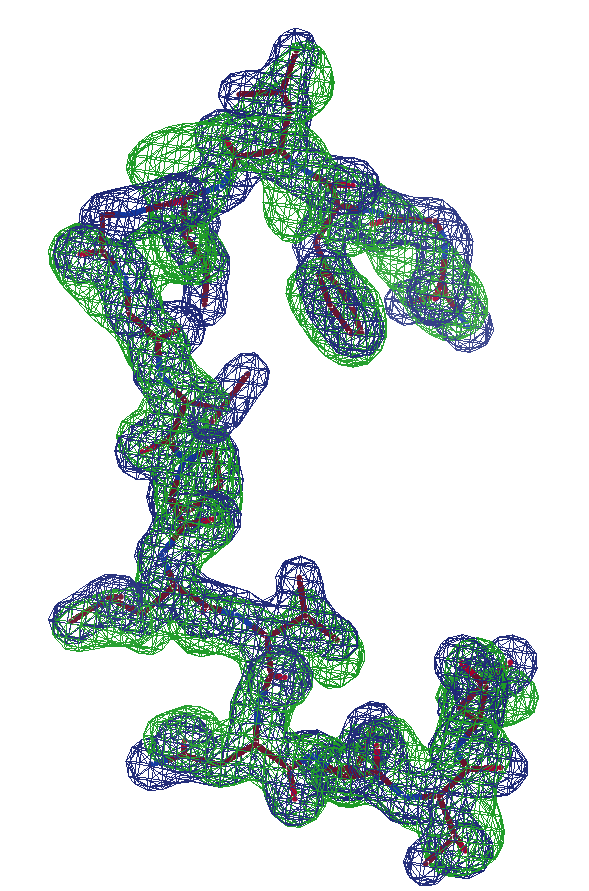}} \vspace{-0.3cm} \hfill
\caption{\small Visualization of two successful predictions after a refining training run; ground truth density maps shown in blue and predictions shown in green.} %The model used to generate the ground truth electron density is again shown in stick representation for reference.}
\label{pred15} \vspace{-0.3cm}
\end{wrapfigure}
%\end{figure}

\begin{figure}[!htp]
\vspace{-0.4cm}
\centering
\subfigure[\empty]{\includegraphics[width=0.44\textwidth]{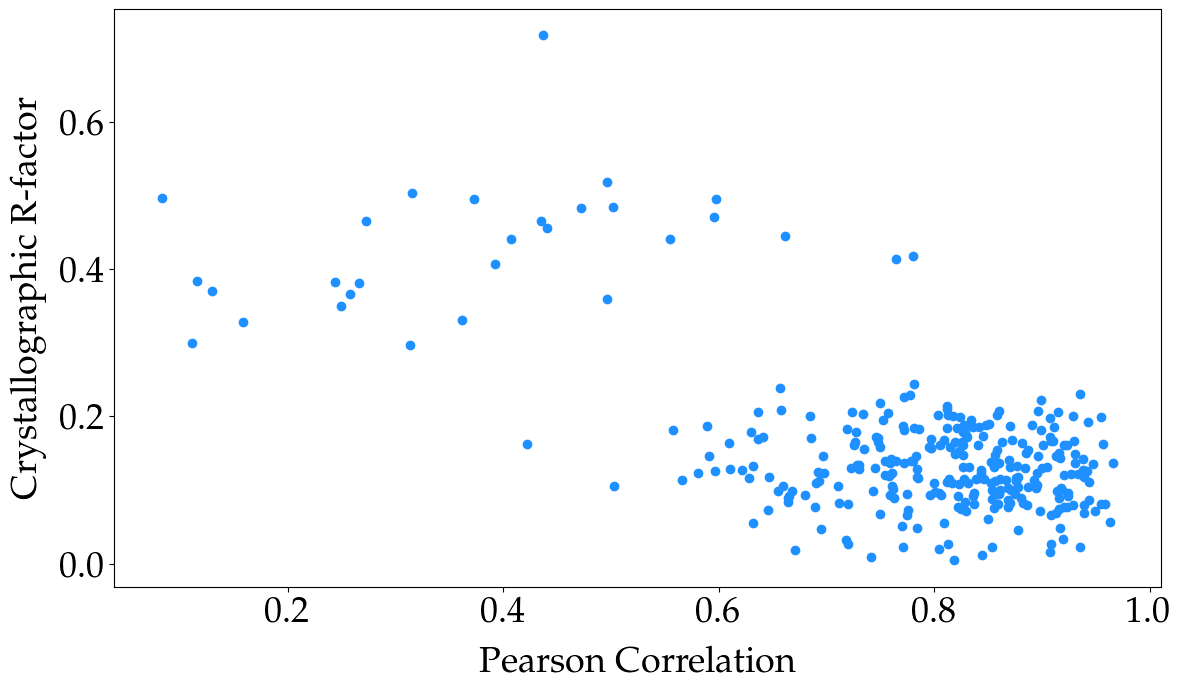}} \hspace{0.5cm}
\subfigure[\empty]{\includegraphics[width=0.44\textwidth]{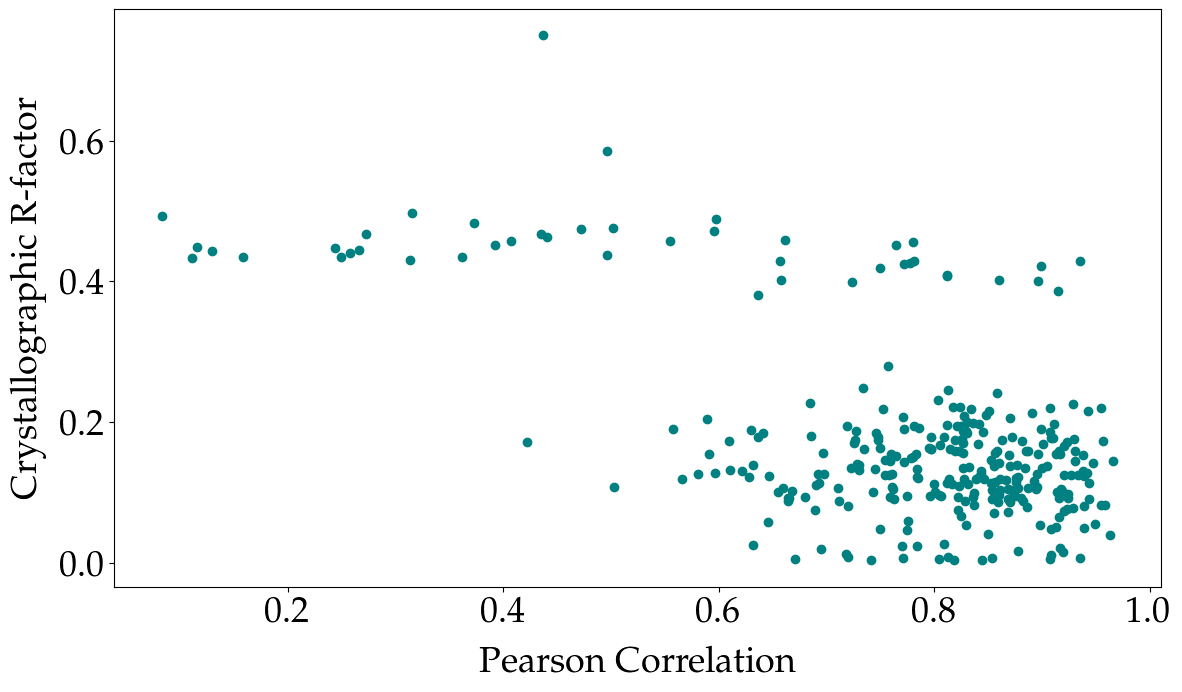}} \hfill
\vspace{-0.3cm}
\caption{\small \textbf{Left Panel}: Scatterplot of post-refinement model R-factors, with solvent flattening applied. 
\textbf{Right Panel}:  Scatterplot of post-refinement model $R$-factors, without solvent flattening applied} 
\label{autobuild15} \vspace{-0.3cm}
\end{figure}

We used the \textit{Autobuild} program within the \textit{PHENIX} suite \citep{Terwilliger2008:ba5109, Liebschner2019:di5033} to perform automated model building and crystallographic refinement on a randomly selected subset of $302$ test set predictions after the refining training run.  We found that $281$ out of $302$ ($\sim 93\%$) refined to a final atomic model with a crystallographic $R$-factor of less than $0.38$, indicating success, when solvent flattening was applied.  Without solvent flattening, $258$ out of $302$ ($\sim 85\%$) refined to such an $R$-factor (performing solvent flattening is known to be especially effective for unit cells with high solvent content, i.e. a large amount of empty space around the atoms).  Figure \ref{autobuild15} shows these results as scatterplots; clearly only a small fraction of the subset of predictions did not refine successfully. And even if no refinement was performed at all, and instead an atomic model was repeatedly fit to our predicted electron densities, we found that $229$ out of $302$ ($\sim 76\%$) of the best such atomic models still had a crystallographic $R$-factor of less than $0.38$. 

Furthermore, after automatic map interpretation using the autobuilding routines in \textit{shelxe} \citep{Uson2018:ba5271} to obtain a poly-alanine chain from each of the $16230$ test set predictions, we found that almost $74\%$ of the resulting models had calculated amplitudes with a Pearson correlation of at least $0.25$ to the true underlying data.  Historical results indicate that further refinement would very likely produce a "correct" model if the initial poly-alanine model has at least such a correlation.

%% file: discussion.tex
\vspace{-0.2cm}\section{Discussion}\label{sec:discussion}
\vspace{-0.2cm}

We have shown that \texttt{CrysFormer} outperforms state of the art models for predicting electron density maps from corresponding Patterson maps in all metrics on a newly introduced dataset (dipeptide). 
Overall, \texttt{CrysFormer} requires fewer epochs to reasonably converge and has a smaller computational footprint. %than state of the art solutions. and only compares unfavorably in terms of memory usage.
Furthermore, our ``refining'' procedure greatly improves training for the vanilla \texttt{U-Net} architecture on our dipeptide dataset, as well as for training \texttt{CrysFormer} on our both dipeptide and 15-residues dataset. 

\textbf{Limitations and next steps.}
Following successful results on our initial 15-residue dataset, we also suggest training our model on variable unit cells at that problem size as future work.  
Eventually, we also prefer handling variable cell angles as well, moving beyond the orthorhombic crystal system.
We will explore changing the formulation of our partial structures to have more than one amino acid residue in a structure, as having each partial structure representing only a single residue may no longer be reasonable, both computationally and from a practical perspective. % in a storage or theoretical sense. 
%Furthermore, we plan to extend the \texttt{CrysFormer} architecture as well: currently, the 3D CNNs utilized in our model are simply single convolutional layers. 
%We find the residual blocks introduced as part of the BigGAN generative architecture \citep{BrockDS19} an interesting research direction.

\textbf{Broader Impacts.}
Solving the crystallographic phase problem for proteins would dramatically reduce the time and expense of determining a new protein structure, especially if there are no close homologs already in the Protein Data Bank. 
There exist some methods that sometimes work under special conditions \cite{cryst8040156}, or that work sometimes but only at very low resolutions \cite{David:gr0256}. 
%None of these are in common use because they are not general or reliable. 
The recent line of work on AlphaFold \cite{jumper2021highly, tunyasuvunakool2021highly} definitely helps in these problems; we note though that this is true mostly in cases where reliable predictions are possible due to strong homologs and/or extensive sequence data. %Moreover, such approaches require substantial computational power, leading to a nontrivial carbon emission footprint.

% In practice, a constant learning rate worked well when training \texttt{CrysFormer} on our dataset.  On more complex datasets, we could look to use other established learning rate schedulers, such as the OneCycle schedule provided by torch \citep{smith2018superconvergence} or the REX framework \citep{chen2022rex}. 

%% file: app_a.tex
\section{Model Architecture of Baseline U-Net model}
Our U-net architecture can be divided into three phases.
The \textbf{\textit{Encoding Phase}} consists of two 7x7x7 convolutional layers.  The first has 25 output channels while the second has 30 output channels, and both are followed by the standard batch normalization and a ReLU activation. Then, a max pooling operation with kernel size 2x2x2 and stride 2 is used to reduce the height, width, and depth dimensions by a factor of 2. 

The \textbf{\textit{Learning Features Phase}} consists of a sequence of 7 residual blocks. Each of these blocks consists of a 7x7x7 convolutional layer with 30 output channels followed by batch normalization and ReLU activation, and then another 30-channel 7x7x7 convolutional layer with batch normalization but no activation. A squeeze and excitation block \cite{hu2017} occurs at this point, applied with the channel dimension reduced by a factor of 2. Afterward, the residual skip connection is applied, followed by another ReLU activation. At the end of this phase, a naive upsampling operation is used to increase the height, width, and depth dimensions by a factor of 2, thus restoring the original dimensions.

The \textbf{\textit{Decoding Phase}} consists of two 5x5x5 convolutional layers. The first has 25 output channels and is followed by batch normalization and a ReLU activation, while the second produces the model predictions and thus has only a single output channel. Since all elements of the target outputs were constrained to be in the range [-1, 1], we apply a final tanh activation function after this layer. 

In all convolutional layers, the input is "same" padded to preserve height, width, and depth dimensionality. Also, the convolutional layers in the encoding and learning features phases are padded using torch's circular padding scheme to account for the periodic nature of the input Patterson maps. Furthermore, all convolutional layers were initialized using the kaiming\_normal function of the default torch.nn module \cite{2015kaiming}. As with the \texttt{CrysFormer}, our U-net model is robust to training batches of examples with differing height, width, and depth shapes.

\section{Additional Details on Dataset Generation}
To start preparing our dataset, we selected nearly 24000 representative Protein Data Bank (PDB) entries using the following criteria:  proteins solved by X-ray crystallography after 1995, sequence length $\geq$ 40, refinement resolution $\leq$ 2.75, refinement R-Free $\leq$ 0.28, with clustering at 30$\%$ sequence identity.  The standardized modifications we applied to each viable coordinate file were as follows: all temperature factors were set to 20, any selenomethionine residues were rebuilt as methionine, and all hydrogen atoms were removed leaving only carbon, nitrogen, oxygen, and potentially sulfur.

In our dataset generation process, an effort was taken to ensure diversity by sampling from PDB entities with low sequence similarity to each other. However, both test and training sets are taking random samples from the conformations allowed in rotamer and Ramachandran space. Any similar conformations would be expected to be in a different rotational orientation in the cell by the nature of the selection process. We did not compute all-versus-all clustering or force the test and training sets to sample distinct conformational regions.  For our 15-residue dataset, in order to obtain a greater amount of starting coordinate files, we allowed at most 3 residues to be shared between distinct examples.  To prevent potential overfitting that could arise from this sharing of subsegments, we enforced that all examples derived from the same initial .pdb file would be placed together in either the training or test set.

Another issue regarding ambiguity in Patterson map interpretation is the fact that an electron density will always have the exact same Patterson map as its corresponding centrosymmetry-related electron density. \cite{hurwitz2020patterson} provided a workaround that involved combining a set of atoms with the set of its centrosymmetry-related atoms into a single example output. However, this also requires a separate post-processing algorithm to separate the original and centrosymmetric densities for each of his model's predictions. Since we are working with real-world structures --rather than randomly placed data-- we can exploit their known properties. In particular, we know that all proteinogenic amino acids are naturally found in only one possible enantiomeric configuration \citep{Helmenstine}. Although the mirror-image symmetry of enantiomers is not exactly the same as centrosymmetry, we show that this is enough to allow us to work with true electron densities of protein fragments. 

\section{Description of Dataset Subset}
Due to limitations of online storage space, we provide a subset of our generated dataset.  This subset represents a total of 200000 dipeptide examples. As expected, patterson.tar.gz contains the generated Patterson maps, while electron\_density.tar.gz contains the corresponding electron densities.  Meanwhile, partial\_structure.tar.gz contains both of the partial structures for each dipeptide example in the subset.

The dataset can be downloaded through this link: 

https://drive.google.com/drive/folders/1X7YkxDd7yTC1RTG1z3NbdRIfKLfFtkrx?usp=share\_link

We will also provide a dataset of prepared .pdb coordinate files of 15-residue examples, to which our dataset generation process can be applied in order to produce Patterson map and electron density tensors.

\section{Additional Visualizations of Model Predictions}

\begin{figure}[!htp]
\vspace{-0.4cm}
\subfigure[\texttt{U-Net+R}]{\includegraphics[width=0.3\textwidth, height= 3.4cm]{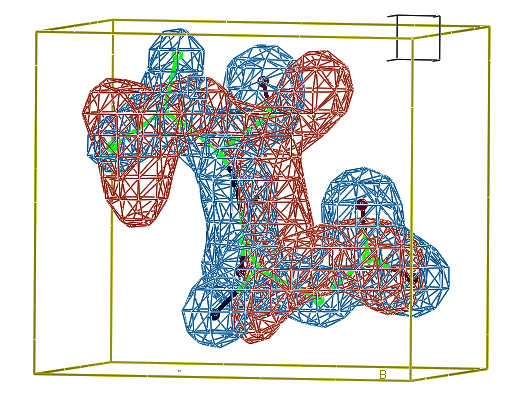}} \hfill
\subfigure[\texttt{U-Net+PS+R}]{\includegraphics[width=0.3\textwidth, height= 3.4cm]{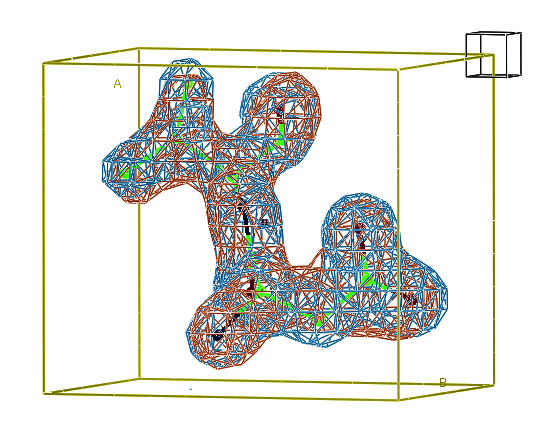}} \hfill
\subfigure[\texttt{CrysFormer}]{\includegraphics[width=0.3\textwidth, height= 3.4cm]{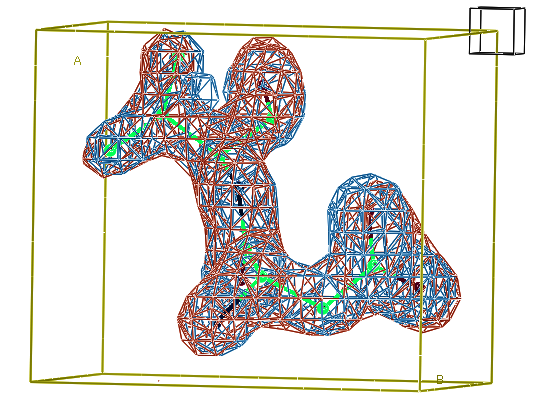}} \hfill
\vspace{-0.35cm}
\caption{Aspartic Acid + Valine} \label{fig:AV}
\subfigure[\texttt{U-Net+R}]{\includegraphics[width=0.3\textwidth, height= 3.5cm]{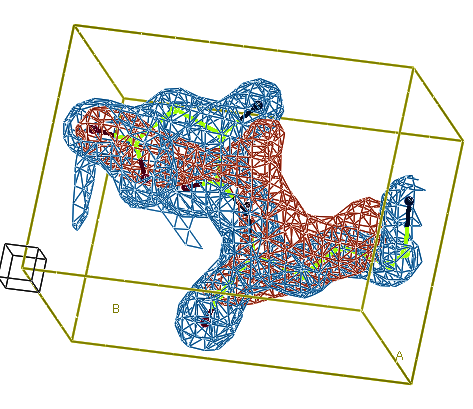}} \hfill
\subfigure[\texttt{U-Net+PS+R}]{\includegraphics[width=0.3\textwidth, height= 3.5cm]{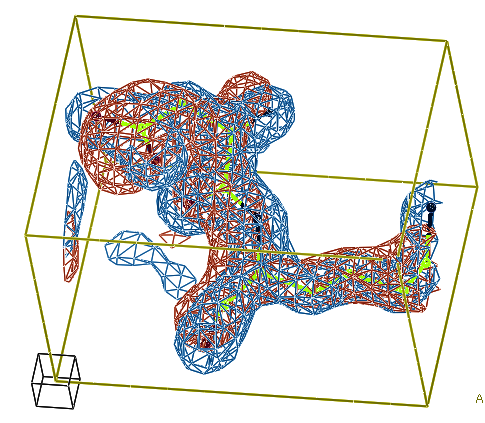}} \hfill
\subfigure[\texttt{CrysFormer}]{\includegraphics[width=0.3\textwidth, height= 3.5cm]{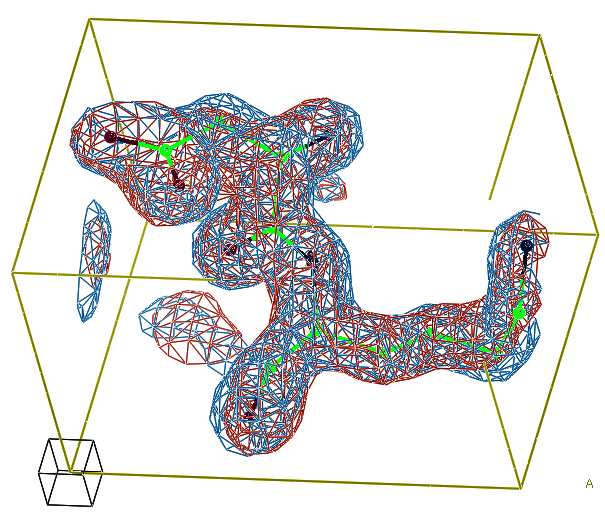}} \hfill
\vspace{-0.35cm}
\caption{Aspartic Acid + Lysine} \label{fig:AL}
\caption{\small Visualizations for dipeptide dataset. Ground truth density maps are shown in blue, while predictions are shown in red. The model used to generate the ground truth electron
density is shown in stick representation for reference.}
\label{fig:mainE1}
\end{figure}

Figure \ref{fig:AV} represents an example in which the additional partial structure input channels provided to the U-Net provided a substantial increase in prediction quality, allowing it to produce a prediction similar to that of the \texttt{CrysFormer}.  
Figure \ref{fig:AL} represents an example in which both providing additional input channels to the U-Net and switching to \texttt{CrysFormer} provided noticeable improvements in prediction quality.

\begin{figure}[!htp]
\vspace{-0.8cm}
\subfigure[{\texttt{4KNK$\_1$.pd$\_$73} CC 0.60 (Rank 11\%)}]{\includegraphics[width=0.49\textwidth, height= 5 cm, keepaspectratio]{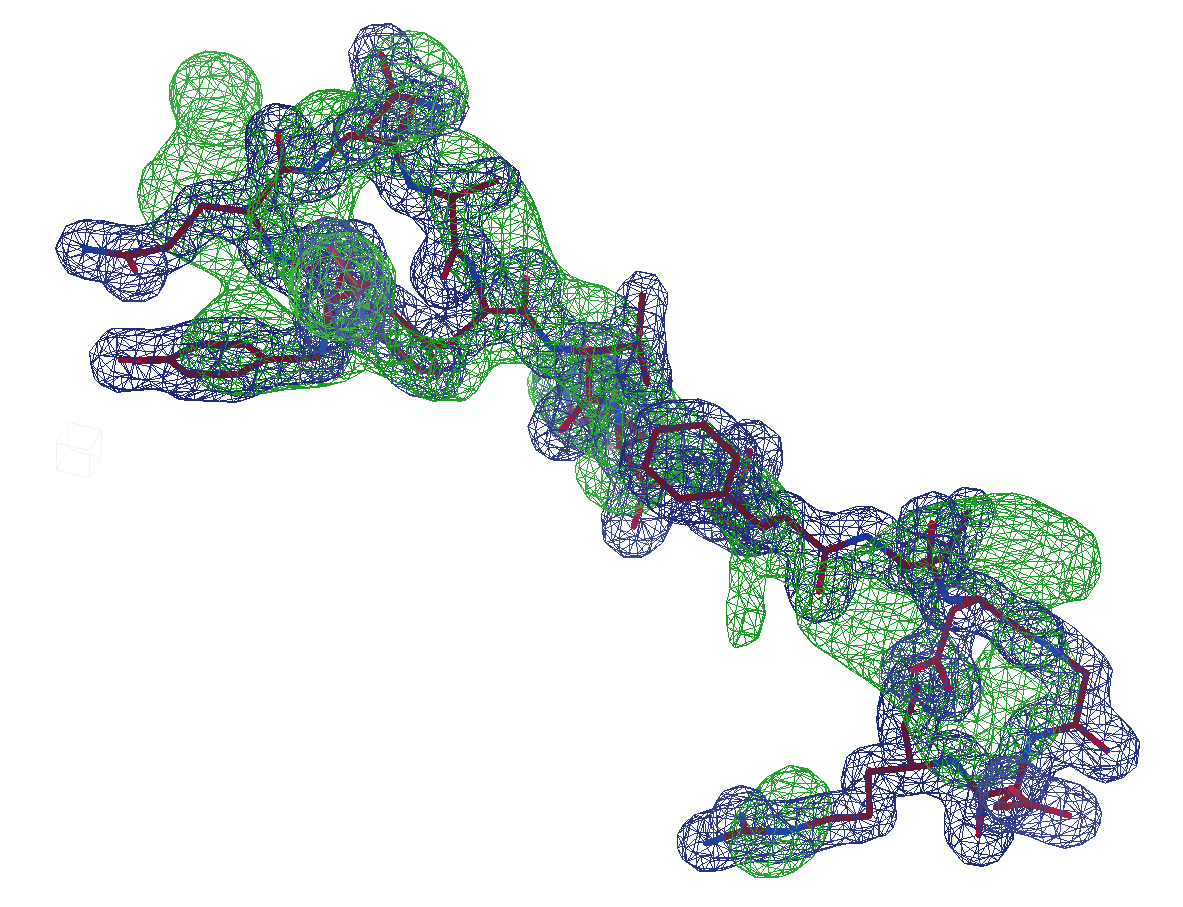}} \hfill
\subfigure[{\texttt{7F1T$\_$1.pd$\_$13} CC 0.66 (Rank 18\%)}]{\includegraphics[width=0.49\textwidth, height= 5 cm, keepaspectratio]{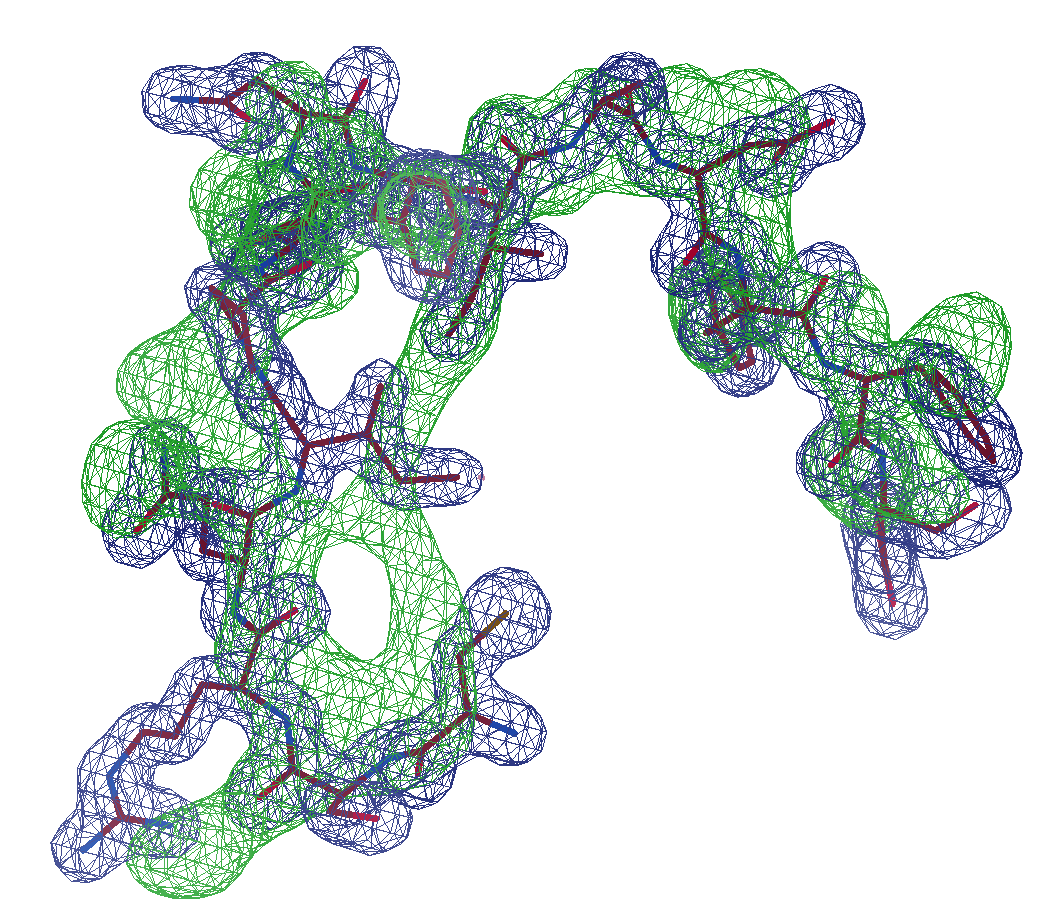}} \hfill
\vspace{-0.2cm}
\subfigure[{\texttt{4XWH$\_$1.pd$\_$380} CC 0.75 (Rank 31\%)}]{\includegraphics[width=0.49\textwidth, height= 5 cm, keepaspectratio]{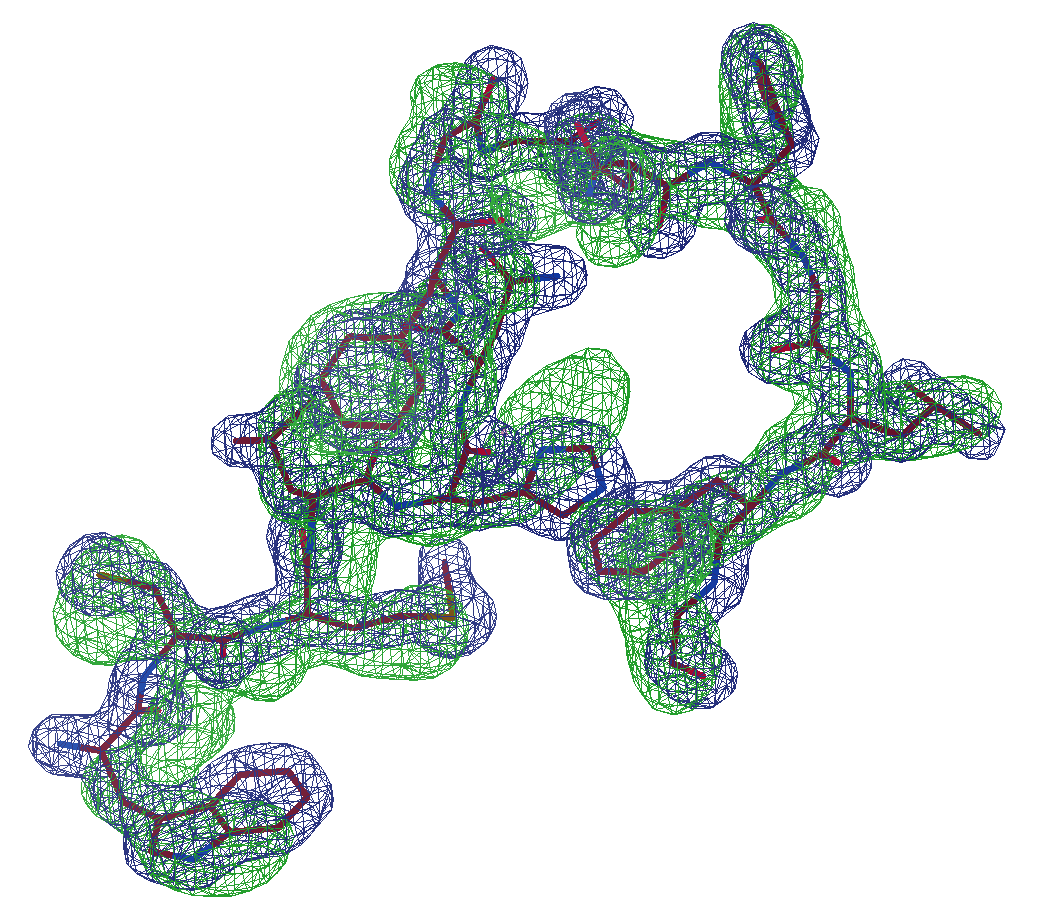}} \hfill
\subfigure[{\texttt{5MSX$\_$1.pd$\_$193} CC 0.76 (Rank 36\%)}]{\includegraphics[width=0.49\textwidth, height= 5 cm, keepaspectratio]{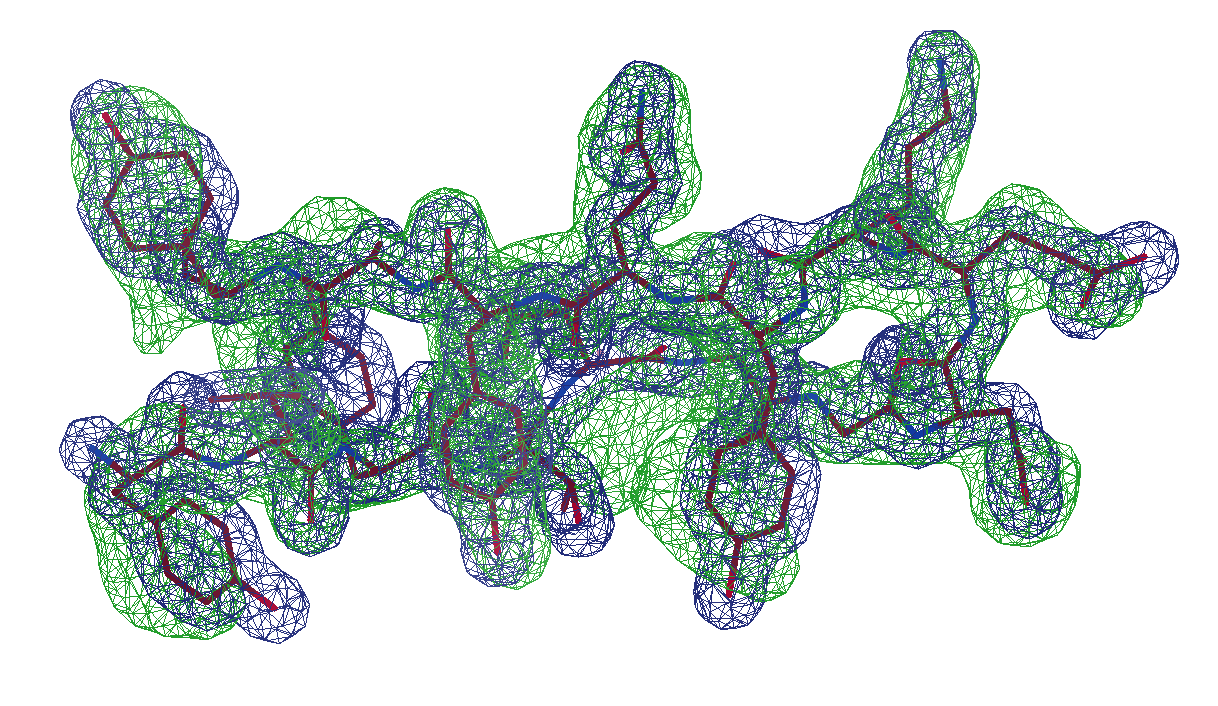}} \hfill
\vspace{-0.2cm}
\subfigure[{\texttt{4FBC$\_$1.pd$\_$121} CC 0.78 (Rank 38\%)}]{\includegraphics[trim=-8cm 0cm -8cm 0cm, clip=true, width=0.49\textwidth, height= 5 cm, keepaspectratio]{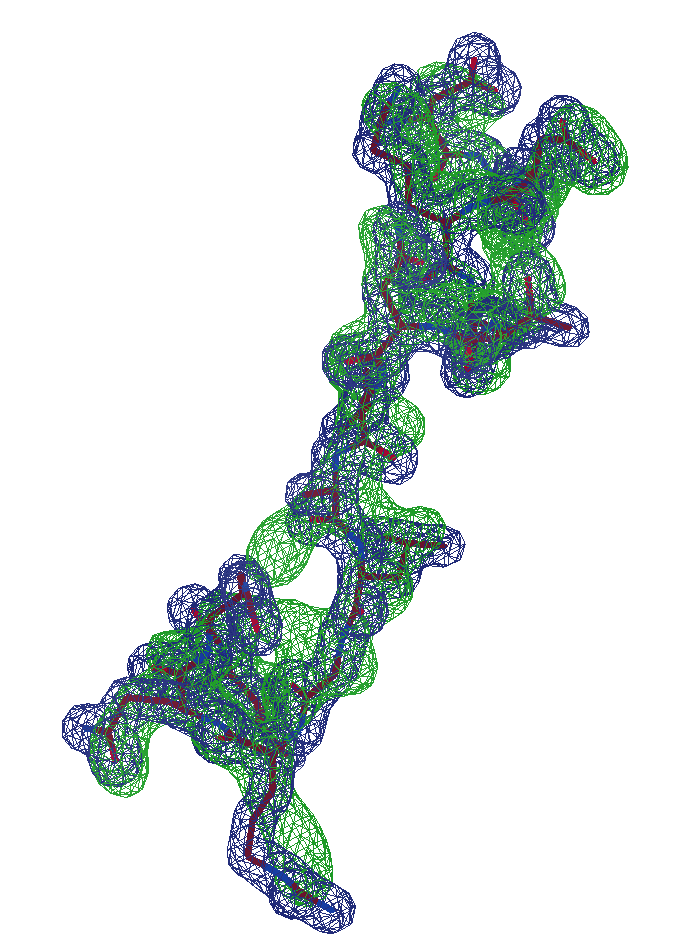}} \hfill
\subfigure[{\texttt{7K34$\_$1.pd$\_$145} CC 0.84 (Rank 57\%)}]{\includegraphics[width=0.49\textwidth, height= 5 cm, keepaspectratio]{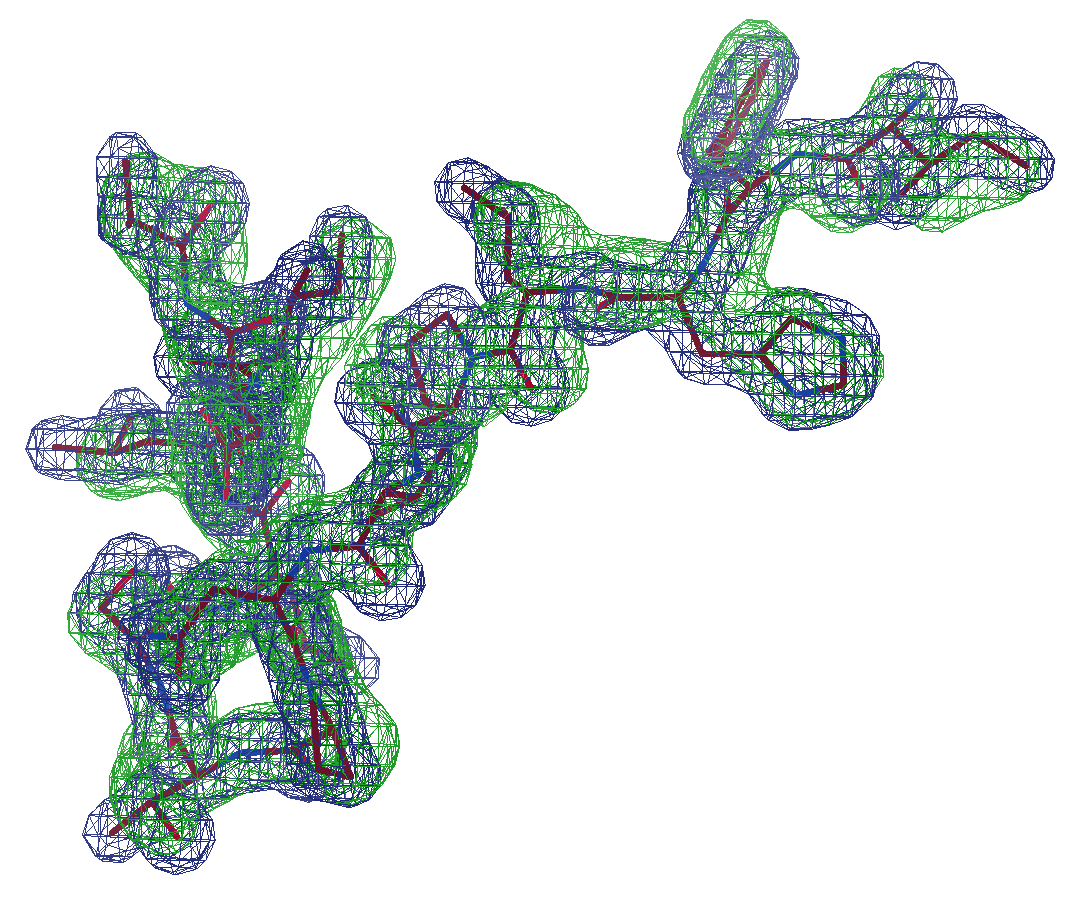}} \hfill
\vspace{-0.2cm}
\subfigure[{\texttt{7F1T$\_$1.pd$\_$13} CC 0.87 (Rank 63\%)}]{\includegraphics[trim=-6cm 0cm -6cm 0cm, clip=true, width=0.49\textwidth, height= 5 cm, keepaspectratio]{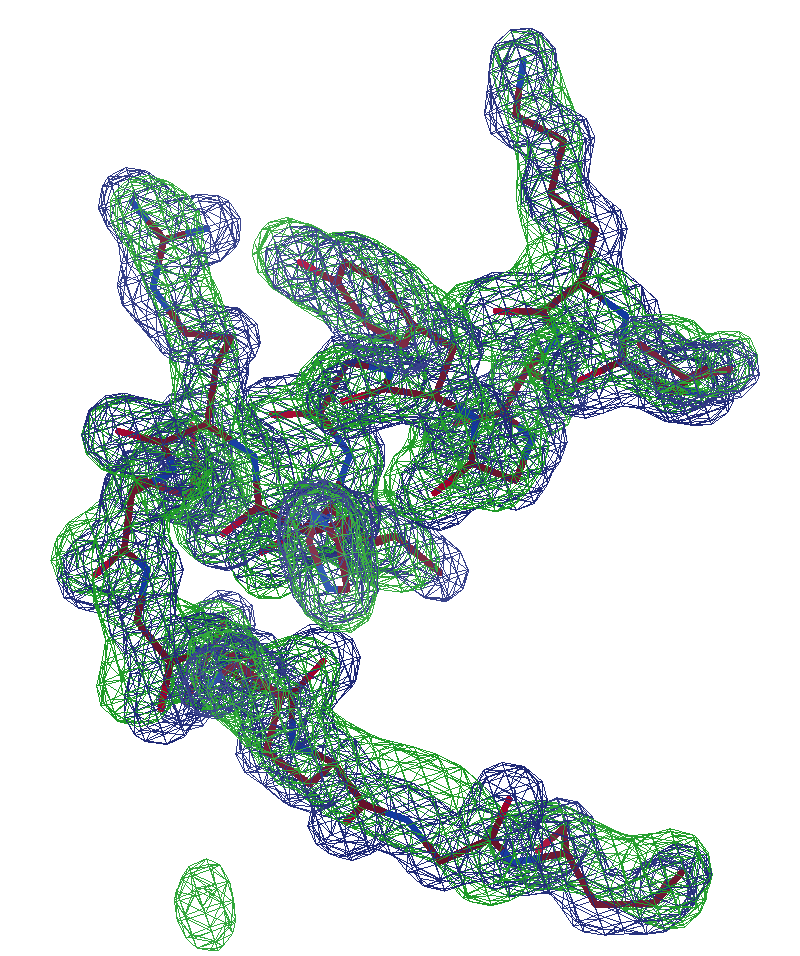}} \hfill
\subfigure[{\texttt{4TXJ$\_$1.pd$\_$37} CC 0.92 (Rank 90\%)}]{\includegraphics[width=0.49\textwidth, height= 5 cm, keepaspectratio]{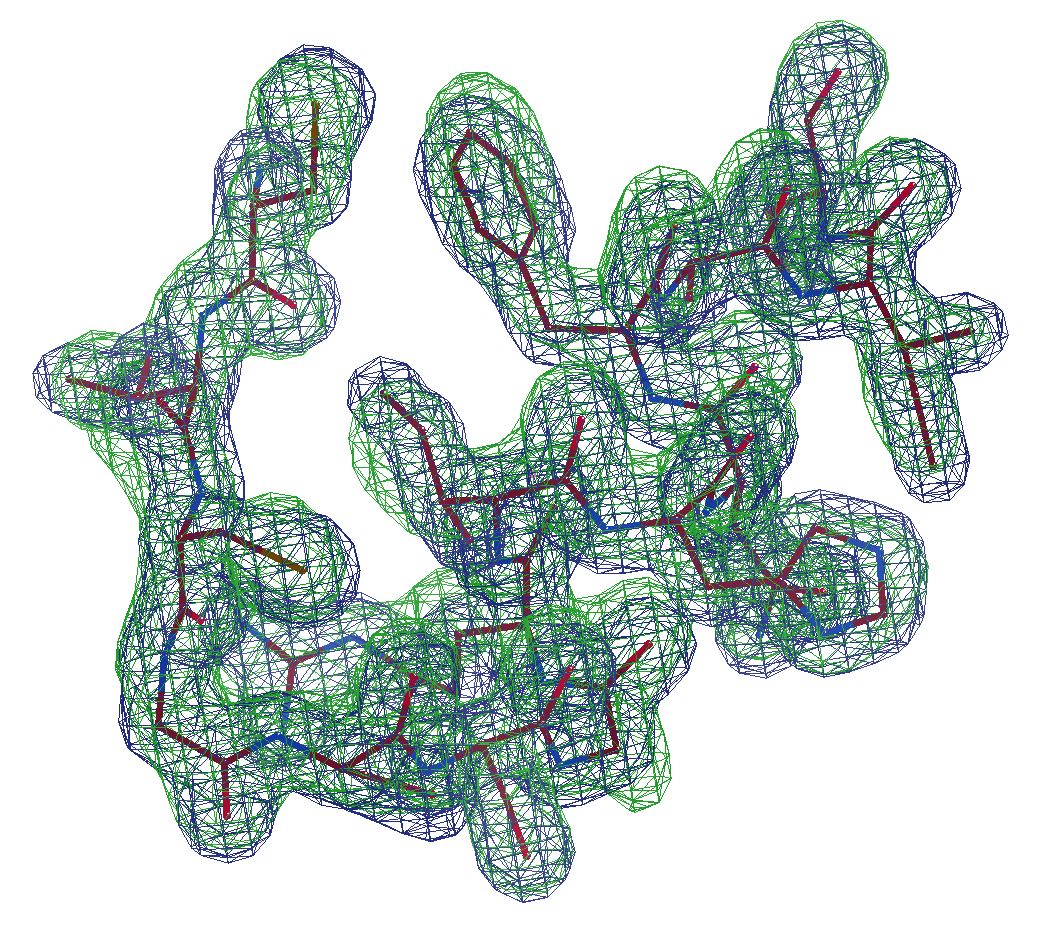}} \hfill
\vspace{-0.2cm}
\caption{\small Visualizations for 15-residue dataset. Ground truth density maps are shown in blue, while predictions are shown in green. The model used to generate the ground truth electron
density is shown in stick representation.}
\label{fig:more15mer}
\end{figure}

It is clear that as prediction quality increases as indicated by reported Pearson correlation, finer details of the true underlying structure are more likely to be accurately reproduced. The predictions in Figure \ref{fig:more15mer} (e), (f), (g), and (h), as well as Figure \ref{pred15} (a) [rank 55\%] and (b) [rank 82\%], were all successfully refined using all of the mentioned autotracing and refinement procedures. But even for relatively poor predictions such as (a) and (b), the rough overall shape can be reproduced even though several portions have clear inaccuracies.

\begin{figure*}[!htp]
    \vspace{-0.0cm}
    \centering
    \includegraphics[width=0.95\textwidth]{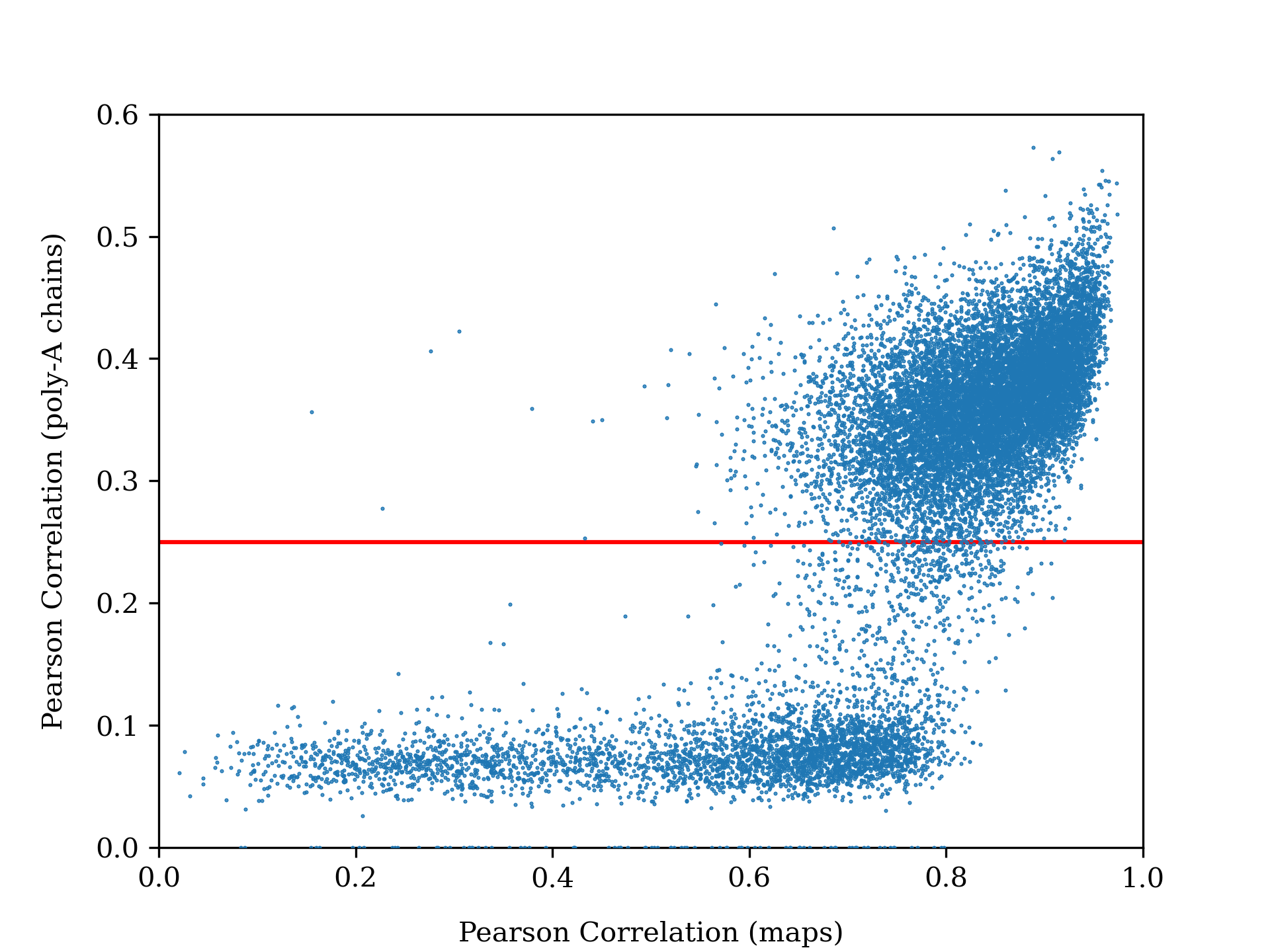} \vspace{-0.1cm}
    \caption{\small Scatterplot of the Pearson correlations of amplitudes of the poly-alanine chains autotraced by \textit{shelxe} to the ground truth amplitudes vs the Pearson correlation of the predicted and ground truth maps for all 16,203 test cases}
    \label{fig:polya}
    \vspace{-0.3cm}
\end{figure*}

Figure \ref{fig:polya} shows the scatterplot of \textit{shelxe} poly-alanine autotracing results on the full 15-residue test set.  As mentioned, examples for which the amplitudes calculated from the initial poly-alanine chain built into the model electron density prediction have a Pearson correlation coefficient with the true underlying structure factor amplitudes of over 0.25 (shown above the red line) are extremely likely to be successfully refined.

%\end{document}